\documentclass[10pt,twocolumn,letterpaper]{article}

\usepackage[pagenumbers]{cvpr} 
\usepackage{graphicx}
\usepackage{amsmath}
\usepackage{amssymb}
\usepackage{booktabs}

\usepackage[pagebackref,breaklinks,colorlinks]{hyperref}

\usepackage[capitalize]{cleveref}
\crefname{section}{Sec.}{Secs.}
\Crefname{section}{Section}{Sections}
\Crefname{table}{Table}{Tables}
\crefname{table}{Tab.}{Tabs.}

\def\cvprPaperID{21} 
\def\confName{EarthVision}
\def\confYear{2023}

\begin{document}

\title{Solar Irradiance Anticipative Transformer}

\author{Thomas M. Mercier\\
Bournemouth University\\
{\tt\small tmercier2@gmail.com}
\and
Tasmiat Rahman\\
University of Southampton\\
{\tt\small t.rahman@soton.ac.uk}
\and
Amin Sabet \\
EscherCloud AI \\
{\tt\small a.sabet@eschercloud.ai }
}
\maketitle

\begin{abstract}
    
This paper proposes an  anticipative transformer-based model for short-term solar irradiance forecasting. Given a sequence of sky images, our proposed vision transformer encodes features of consecutive images, feeding into a transformer decoder to predict irradiance values associated with future unseen sky images. We show that our model effectively learns to attend only to relevant features in images in order to forecast irradiance. Moreover, the proposed anticipative transformer captures long-range dependencies between sky images to achieve a forecasting skill of 21.45~\% on a 15 minute ahead prediction for a newly introduced dataset of all-sky images when compared to a smart persistence model.
\end{abstract}

\section{Introduction}
\label{sec:intro}

Solar energy has emerged as one of the most promising alternatives to non-renewable energy sources. As the photovoltaic (PV) industry grows at pace from gigawatt to terawatt scale, the need for more accurate and efficient forecasting of PV output becomes ever more critical. Grid scale solar based power generation poses challenges for grid operators due to the intermittent nature of the supply\cite{barnesModellingPVClouding2014,linRecentAdvancesIntrahour2022}. Since solar irradiance is a key predictor of PV output, irradiance forecasting on a sub-hour level can greatly support stable and economical power generation. Even forecasting 5 minutes into the future is critical in PV systems to balance storage and load for intermittency as well as having benefits in energy minute by minute trading.
The level of solar irradiance seen in a particular location varies based on the cyclical changes of the season, the sun position throughout the day and the weather conditions. While the first two factors are consistently predictable, weather conditions, especially the level of cloud cover make purely time based predictions inaccurate\cite{rajagukgukReviewDeepLearning2020}.

\begin{figure}[h]
	\centering
	\includegraphics[width=0.7\linewidth]{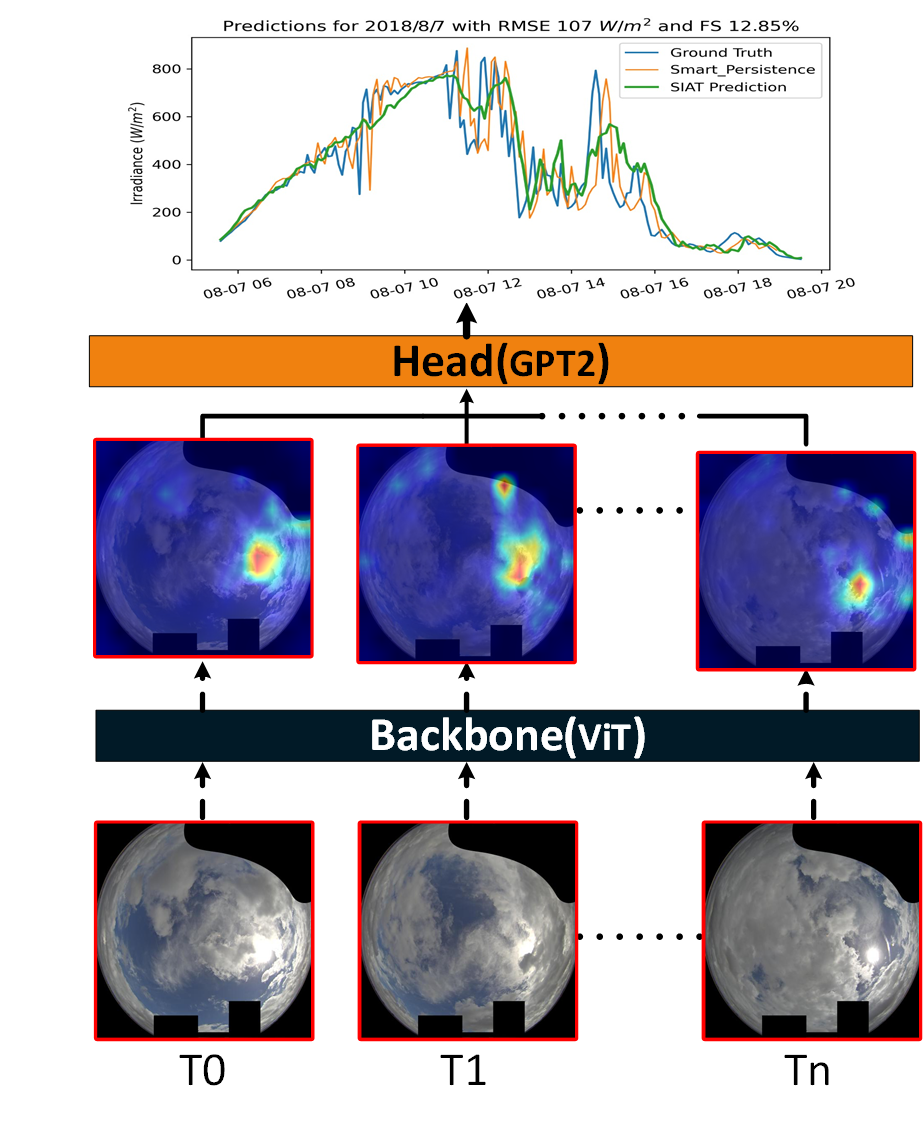}
	\caption{High level overview of model operation. The backbone encodes features from each sky image and the head predicts future irradiance.}
	\label{fig:avt_overview}
\end{figure}

Two common approaches for short term irradiance forecasting are the use of statistical methods derived from past irradiance measurements and image based forecasts using either ground based sky images or satellite imagery\cite{diagneReviewSolarIrradiance2013}. Common deep learning (DL) based approaches make use of convolutional neural networks (CNNs) to extract features from images that can then be used to give an associated irradiance value\cite{linRecentAdvancesIntrahour2022}. Ordinarily to predict irradiance, a series of consecutive images are used in either a 3 dimensional CNN or a combination of a CNN and a long short term memory (LSTM) based architecture\cite{palettaBenchmarkingDeepLearning2021}. This is ultimately based on the temporal information contained in the series of images. In contrast to LSTM based models, transformers offer both the ability to process sequences in parallel as well as excellent modeling of long-term dependencies\cite{khanTransformersVisionSurvey2021}. The recent application of the self-attention based transformer architecture to computer vision tasks combined with the high performance of transformer-based networks for tasks where long term dependencies are crucial, makes this type of network attractive for solar irradiance forecasting\cite{linRecentAdvancesIntrahour2022,vaswaniAttentionAllYou2017}. 
We propose utilizing a self-attention based backbone network that creates feature representations for each frame in a sequence of all-sky images and then using a Generative Pre-trained Transformer 2 (GPT-2) based decoder on the resulting sequence of encoded feature vectors to produce solar irradiance predictions\cite{radfordLanguageModelsAre2018,dosovitskiyImageWorth16x162021}. We refer to our model as Solar Irradiance Anticipative Transformer (SIAT). A high level overview of our approach is depicted in \cref{fig:avt_overview}.

Our contributions are 1) introduction of a purely attention based forecasting framework that only uses images without any auxiliary data and outperforms previous models on three timestep prediction task, 2) evaluation of our model on three datasets and 3) introduction of three stage training procedure and multiple loss components supervision scheme for strong supervision signal.

\section{Related Work}
\label{sec:relwork}

Due to the importance of solar irradiance forecasting, many different approaches have been reported in the literature with classical irradiance modelling being based on meteorological input data such as humidity, rainfall and temperature\cite{kumariDeepLearningModels2021}. The general success and the increasingly low barrier of entry to machine learning, it has seen broad adoption in the physical sciences\cite{daiAccurateInverseDesign2021,liuGenerativeModelInverse2018a}. DL based approaches have become increasingly popular for tackling previously intractable or poorly addressed problems. 

For computer vision based irradiance forecasting approaches a range of different datasets are used in literature. Irradiance predictions are commonly made based on a sequence of past sky images, often combined with auxiliary data or input from a classical prediction model. Typically, a dataset consists of a large collection of all-sky images that can be temporally aligned with irradiance values collected at the same site. The National Renewable Energy Research (NREL) dataset was collected in the state of Colorado in the USA and it is publicly available\cite{stoffelNRELSolarRadiation1981}. The newly introduced Chilbolton dataset was collected in a south England based location and is available upon request\cite{scienceandtechnologyfacilitiescouncilChilboltonFacilityAtmospheric2016,scienceandtechnologyfacilitiescouncilChilboltonFacilityAtmospheric2003}. The SIRTA dataset was collected by the SIRTA Atmospheric Research Observatory, a meteorological institute near Paris in France and the institute makes the dataset available upon request\cite{haeffelinSIRTAGroundbasedAtmospheric2005}. The EDF dataset was collected on La Reunion Island and is not publicly available as it was collected by a private company\cite{guenDeepPhysicalModel2020}.

Le Guen et al. have shown that a time series of all-sky image data in combination with past irradiance data can be used to predict 5 minutes of irradiance data given 5 minutes worth of past data\cite{guenDeepPhysicalModel2020}. The images in their dataset were spaced only one minute apart, offering dense temporal information about changes in sky condition. Their dataset was collected in-house at an EDF test site on La Reunion Island. Their model consists of two sub-models, a convolutional LSTM and PhyDNet, which uses partial differential equations for video prediction tasks. The output of the sub-models is combined to produce an irradiance prediction as well as a sky image prediction. They utilise a very large dataset of 6 million images at a size of 80 by 80 pixels and achieve a nRMSE of 23.5~\% for a 5 minute ahead irradiance forecast.

Wen et al. show that solar forecasting can be achieved without using a sequential model\cite{wenDeepLearningBased2021}. They utilise a ResNet18 architecture with the red channel of the past images stacked as input to their network. On the NREL dataset and another California based dataset they report a forecasting skill (FS) up to 17.7~\% for a 10 minute ahead prediction compared to a smart persistence (SP) model. Please see \cref{sec:modeleval} for details on how the FS is calculated from the SP model.

Gao and Liu utilise a vision transformer (ViT) to encode the information contained in sky images from two NREL datasets as well as auxiliary meteorological data\cite{gaoShorttermSolarIrradiance2022}. A sequence of encoded images and auxiliary information is then fed into another transformer encoder together with a learnable embedding. The output of this encoder is then concatenated with the prediction from a clear sky model and fed into an MLP with residual connections to produce the irradiance forecast. Using one hour worth of past images they report a normalized absolute percentage error of 22.6~\% for a one hour forecast.

Paletta et al. present a benchmarking study of different DL based models with the convolutional LSTM giving the best results\cite{palettaBenchmarkingDeepLearning2021}. All presented models take as an input either a sequence of images or a single image pair, the latter consisting of all-sky images taken at the same time but with different exposure settings. They report a RMSE based FS of 20.4~\% for their best model with a SP model used as a comparison. The same group further improved their predictions by implementing the ECLIPSE, a model that has both irradiance and image segmentation as an output \cite{palettaECLIPSEEnvisioningCLoud2022}. For both studies they utilise a dataset collected and provided by SIRTA laboratory in France which contains all-sky images captured every 2 minutes at 2 different exposure levels\cite{haeffelinSIRTAGroundbasedAtmospheric2005}. They achieve RMSE of 83.8, 98.5, 109.1 $W/m^2$ which corresponded to a RMSE based FS of 8.7, 23.7 and 24.8~\% for 2, 6 and 10 minute ahead irradiance prediction respectively. Since the authors of the ECLIPSE model report that they outperform previous studies we compare our model's performance to this method.

From the presented reports in the literature it is clear that a large variety of prediction approaches exist but that there is a lack of standardisation in the reporting of prediction results as well as in the chosen prediction time horizon. This makes direct performance comparisons difficult. Comparisons are further complicated by the fact that the locations of data collection vary significantly and thus differences in local weather patterns result in datasets of varying difficulty. We will therefore present multiple different performance metrics and evaluate both our model as well as the competing ECLIPSE model on three datasets. Most of the reported works utilize LSTM based networks, which struggle with longer term dependencies. We address this issue by relying on a self-attention mechanism.

\section{Proposed Framework}
\label{sec:methods}
\begin{figure}[t]
	\centering
	\includegraphics[width=0.99\linewidth]{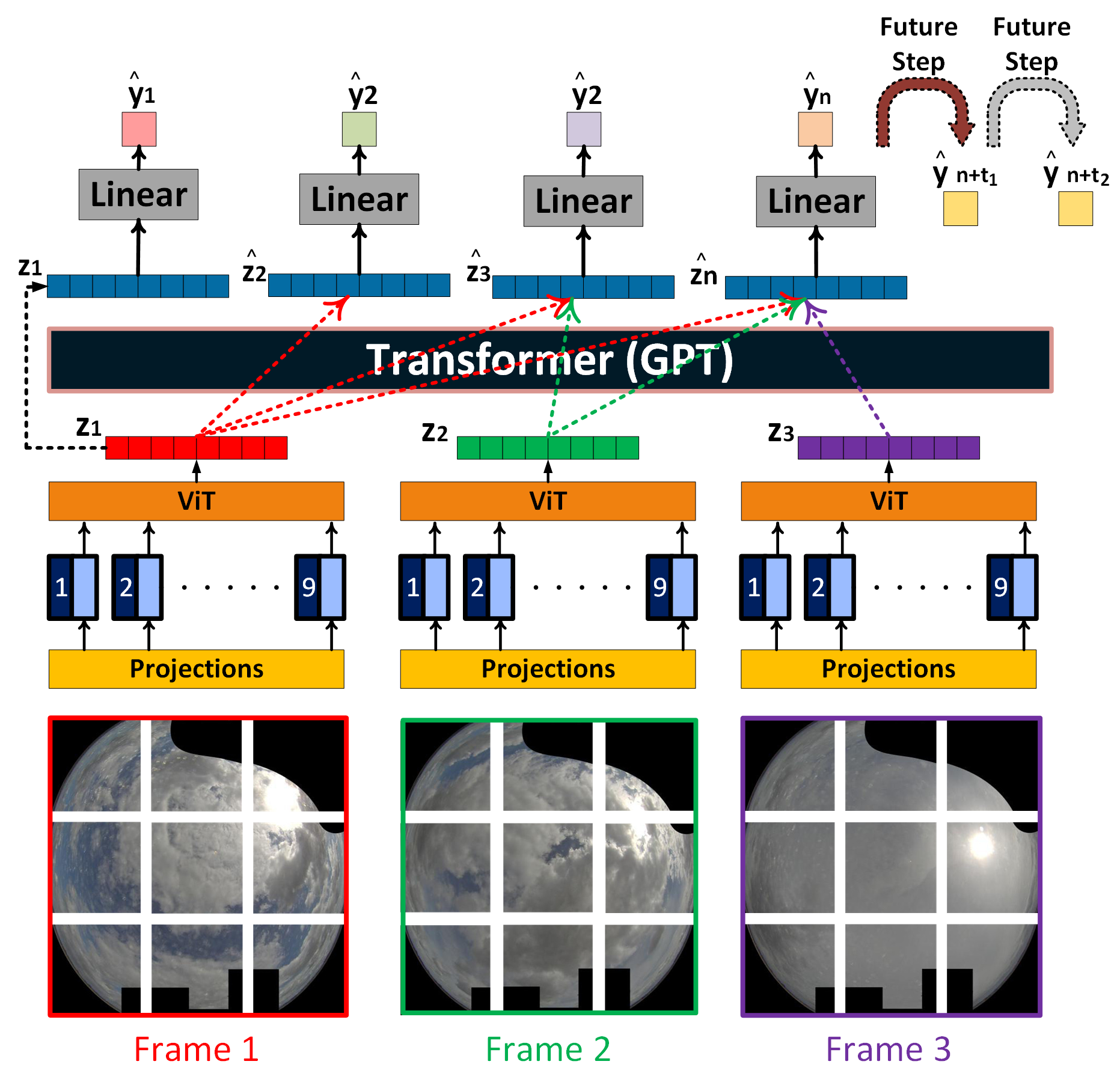}
	\caption{Model flow for SIAT. The ViT backbone encodes the projected flattened image patches into feature vectors $z$ for each image in the input sequence. Together with temporal positional embeddings the feature vectors $z$ are fed into the GPT-2 based decoder. The decoder produces future feature vectors $\hat{z}$ from which irradiance values $\hat{y}$ are produced through a linear layer. For illustrative purposes only 9 image patches and only 3 images are shown as the past context for the model. For visualisation purposes time steps are unfolded; otherwise the same set of weights are used for the ViT backbone and projections to process frames.}
	\label{fig:model_flow}
\end{figure}

\cref{fig:model_flow} depicts the proposed model architecture. Inspired by anticipative video transformer\cite{girdharAnticipativeVideoTransformer2021}, our model utilises a ViT as a backbone \( \mathcal{B} \) which operates on linearly projected flattened image patches $x_t$ to produce an encoding $z_t$ for each of the $s$ input images in the sequence\cite{touvronTrainingDataefficientImage2021}. The input images are split into 16 by 16 patches which are flattened and linearly projected. A class token is prepended to the patch features and a spatial position embedding is added. The output associated with this class token is then used as the image feature representation $z_t$.

\begin{equation}
z_t, z_{t+1},... = \mathcal{B}(x_t),\mathcal{B}(x_{t+1}),...
\end{equation}

From each input image in the sequence, a feature representation is extracted. Together with temporal positional embeddings, this sequence of features is then used by the GPT-2 based decoder \( \mathcal{D} \)\cite{radfordLanguageModelsAre2018}. The decoder consists of four layers of masked multi-head attention, a layer norm and a multi layer perceptron (MLP). The decoder produces one $\hat{z}$ for each timestep in the sequence of $s$ images, which are then put through a linear layer \( \mathcal{L} \) to produce an irradiance value $\hat{y}_{t+1}$. The linear layer is a fully connected layer with the number of input neurons depending on the dimensionality of the $\hat{z}$ and the output being a single value, representing the predicted irradiance.
\begin{align}
\hat{z}_{t+1},...,\hat{z}_{t+s+1} &= \mathcal{D}(z_t,...,z_{t+s})\\
\hat{y}_{t+1} &= \mathcal{L}(\hat{z}_{t+1})
\end{align}

$\hat{z}_{t+1}$ here represents the predicted image features one timestep ahead of the past image feature $z_t$. The masked attention of the GPT-2 decoder ensures that the model can only attend to past features to make the prediction. To predict multiple timesteps into the future, the predicted feature vector is appended to the past context and this is then fed into the head decoder network to predict another step into the future.
The proposed framework is both purely attentional in nature and purely image based with no auxiliary data such as past irradiance values, cloud cover or sun location being utilized to make the irradiance predictions. This significantly reduces the requirements for deployment as the equipment needed to collect such auxillary data can present a significant expense.

\subsection{Training}
As \cref{fig:model_flow} shows, the GPT-2 decoder produces predicted image features which are then fed through a linear layer to give an irradiance prediction. During training of the full model the presented architecture allows for optimization using two loss metrics. 

\begin{align}
L_{irr} &= \frac{1}{n} \sum \left( y_t - \hat{y}_t \right)^2\\
L_{enc} &= \frac{1}{n} \sum \left( z_{t} - \hat{z}_{t} \right)^2
\end{align}

$L_{irr}$ represents the difference between the predicted irradiance $\hat{y}_{t}$ and ground truth irradiance $y_t$, and $L_{enc}$ the difference between the encodings $z_t$ and $\hat{z}_{t}$. $L_{irr}$ can be further separated into the loss associated with the intermediate irradiance predictions and the final prediction, the latter of which is ultimately what is of interest. 

The GPT-2 based decoder utilizes masked multi-head attention and hence only attends to encoded features before the time of the prediction. This allows the model to simultaneously predict irradiance values for all input timesteps. Thus an input sequence of four images will produce five irradiance values with all but the first irradiance resulting from the decoder output. If more than one future timestep is to be predicted, the model can be unrolled to predict future image encodings by iteratively adding the intermediate predictions to the past context. Supervising the difference between $\hat{z}_{t}$ and $\hat{z}_{t+1}$ ensures that the decoder is able to predict future encoded features. During training the model supervision is based on a weighted sum of both loss $L_{irr}$ and $L_{enc}$ as follows.

\begin{equation}
\label{eq:loss_comp}
L_{total} = \alpha L_{irr,f} + \beta L_{irr,i} + \gamma L_{enc}
\end{equation}

Here, $\alpha$, $\beta$ and $\gamma$ represent the weight of the loss associated with the final ($L_{irr,f}$) and intermediate ($L_{irr,i}$) irradiance predictions and the encoding ($L_{enc}$) prediction, respectively. We train the model in three stages. The first stage consists of training the backbone ViT to map a single image to a single irradiance value. During this training stage the ViT is only supervised by $L_{irr}$. During the second stage of training, the mapping trained ViT model has its regression head removed and the remaining model is frozen and used as the image encoding backbone of the overall architecture. With this frozen backbone, the head GPT-2 based decoder is then trained to predict future encodings which are turned into irradiance value predictions via a linear head. During this and the following stage all loss components are used to supervise the model. In the third stage the backbone model is unfrozen and the model is fine-tuned back to back.

\subsection{Model evaluation}
\label{sec:modeleval}

Since every solar irradiance model ultimately aims to give an accurate prediction of a continuous value, error metrics commonly used for regression tasks such as mean absolute error (MAE), mean squared error (MSE) and RMSE can be employed. However, since irradiance values are strongly weather dependent the mean and variance of a given dataset can vary substantially for different measurement locations. A region with largely clear skies will produce irradiance values that vary smoothly over time and are therefore much easier to predict. A simple difference based error metric would not take the differences in prediction difficulty into account. To improve comparability of model performance on different datasets, evaluation metrics can be normalised by dividing them by the mean of the training irradiance values\cite{guenDeepPhysicalModel2020}. A normalised RMSE will be abbreviated by nRMSE. While this improves comparability, it is generally recommended to use a FS metric that compares the error achieved by the presented model to the error achieved by a reference model \cite{yangVerificationDeterministicSolar2020a}. An overview of the loss metrics is given below.

\begin{align}
MAE &= \frac{1}{n} \sum \left| y - \hat{y} \right|\\
RMSE &= \sqrt{\frac{1}{n} \sum \left( y - \hat{y} \right)^2}\\
FS &= 1 - \frac{RMSE_{model}}{RMSE_{reference}}
\end{align}

A FS above 0 indicates that the model in question outperforms the reference model. The FS can be calculated based on any loss metric that can be computed for both models. However, the RMSE based FS is the most commonly used metric. Clear sky irradiance and SP are the most commonly used reference models \cite{yangVerificationDeterministicSolar2020a,inmanSolarForecastingMethods2013,inmanImpactLocalBroadband2015}. Clear sky irradiance models use meteorological data such as aerosol optical density and air pressure in combination with the location and time of year to model what the irradiance would be without cloud cover. SP models use the most recent observation in the data as a prediction with the value being adjusted by a clear sky index, as shown in \cref{eq:fs_eq}. This index can either be derived from a clear sky irradiance model or be based on measured data.

\begin{equation}
\label{eq:fs_eq}
\hat{y}_{t+T} = \frac{y_t}{y_{clear,t}} y_{clear,t+T}
\end{equation}

Here, $\hat{y}_{t+T}$ represents the predicted irradiance at time $t+T$, $y_t$ the real irradiance value at time $t$ and $y_{clear}$ the clear sky model prediction for time $t+T$. We use the simplified Solis model to calculate the clear sky index needed for the SP reference model\cite{ineichenBroadbandSimplifiedVersion2008,ineichenValidationModelsThat2016}. The Solis model requires meteorological data such as air pressure, aerosol optical depth and precipitable water as input, which is sourced from\cite{gilesAdvancementsAerosolRobotic2019,eckClimatologicalAspectsOptical2010,eckFogCloudinducedAerosol2012,eckColumnarAerosolOptical2005,scienceandtechnologyfacilitiescouncilChilboltonFacilityAtmospheric2003a}. Since the SP model's predictions simply shift the ground truth irradiance by a multiple of the timestep (5 minutes in the case of the Chilbolton dataset) with a small adjustment based on a clear sky index, the SP model's prediction appear to follow the ground truth relatively well for very short term predictions. However, this method still results in a large average error as is illustrated in \cref{fig:Hist_joint_Chilbolton_Smart_Persistence}. The FS expresses how much a model outperforms this approach.

\section{Datasets}

\begin{figure}[h]
	\centering
	\begin{subfigure}{0.39\linewidth}
		\centering
		\includegraphics[width=0.99\linewidth]{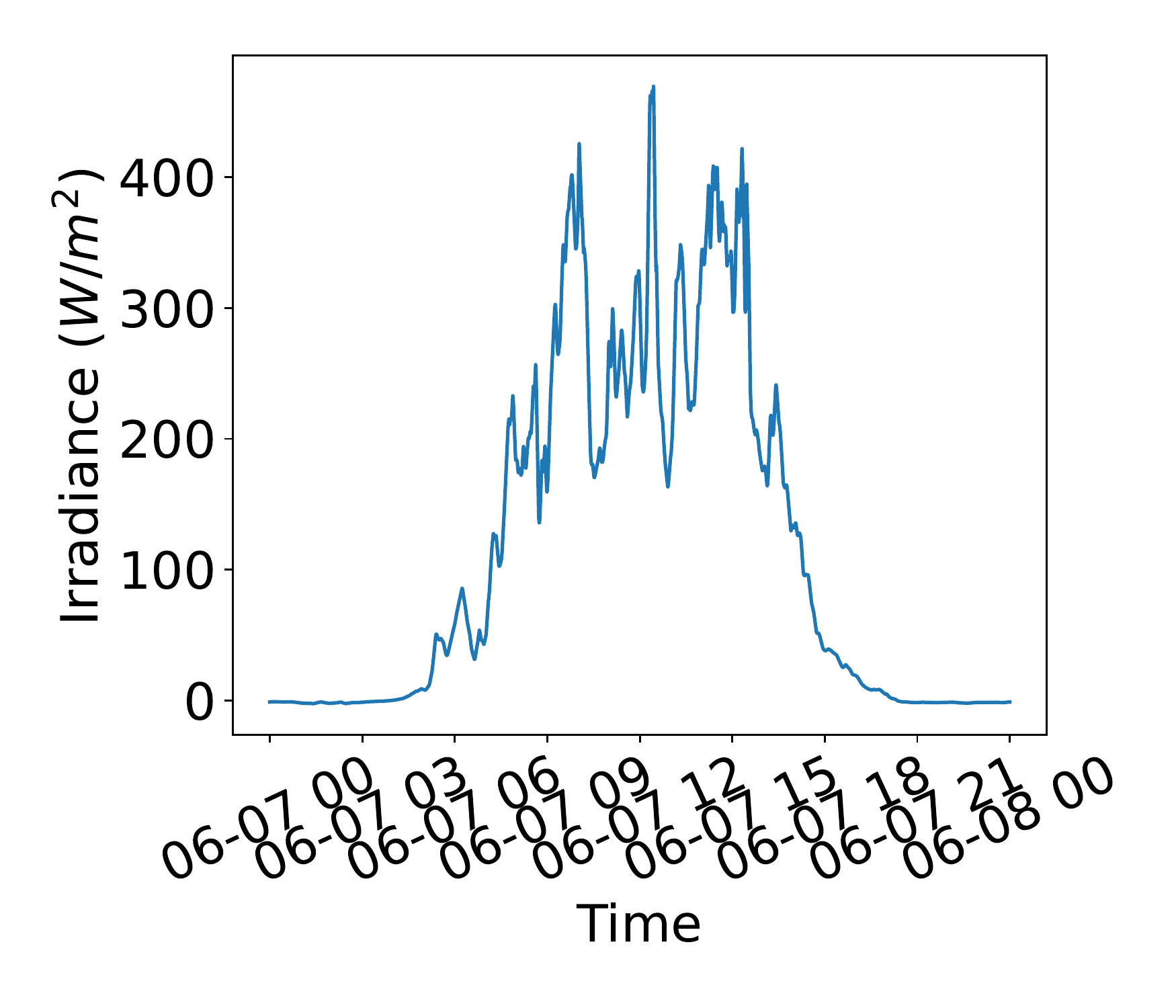}
		\caption{Unprocessed irradiance.}
		\label{fig:raw_RVC}
	\end{subfigure}
	\begin{subfigure}{0.39\linewidth}
		\centering
		\includegraphics[width=0.99\linewidth]{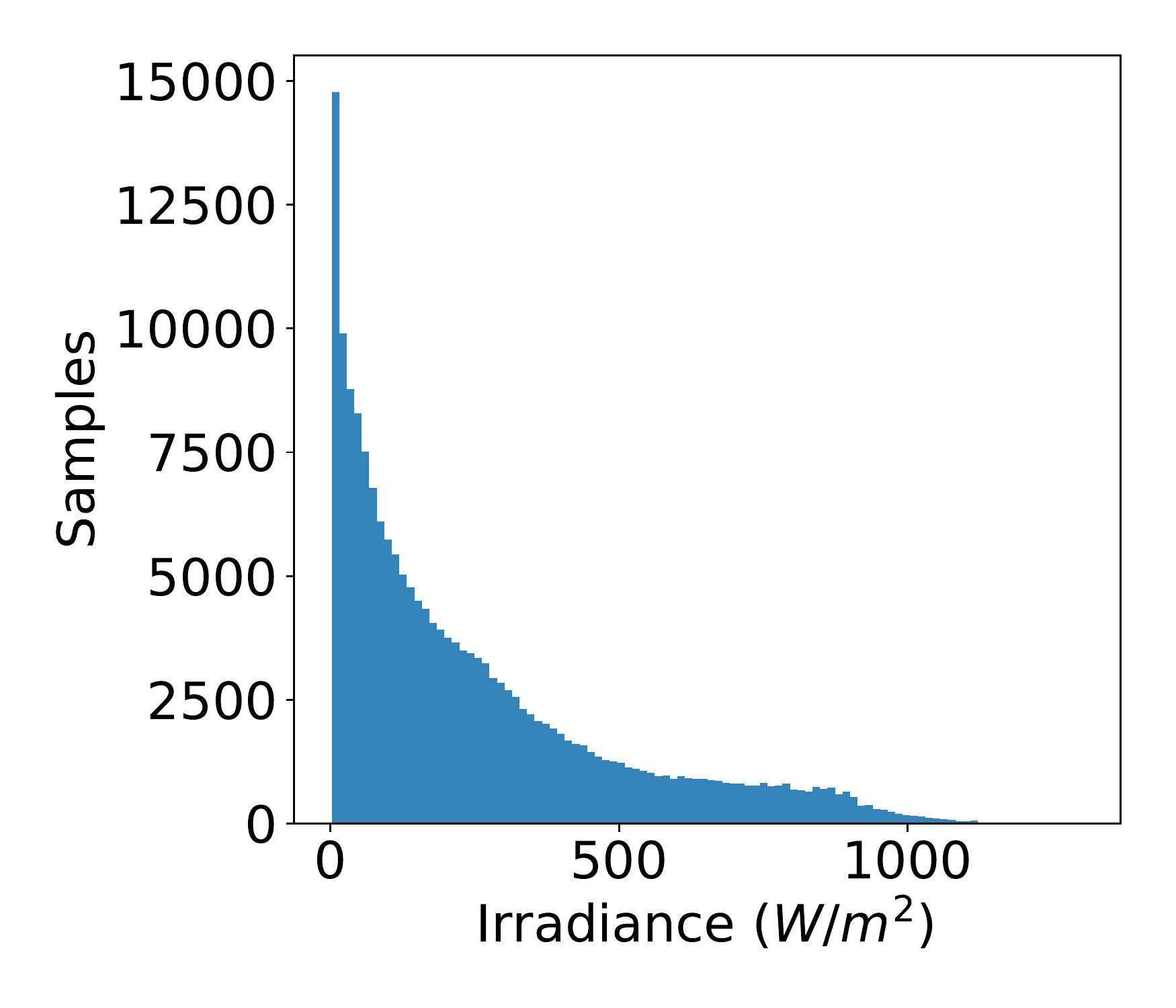}
		\caption{Histogram for irradiance values.}
		\label{fig:RVC_Hist_Threshold_10_Normal_Images}
	\end{subfigure}
	\caption{Raw irradiance data for an example day as well as the distributions of values after pre-processing and filtering for the Chilbolton dataset.}
	\label{fig:raw_processed_data_overview}
\end{figure} 

In addition to evaluating our SIAT model's performance on two datasets previously used in literature, we introduce the new Chilbolton dataset. In contrast to previously used datasets the Chilbolton dataset was collected in the challenging weather patterns of the south of the UK. The sky images were provided by the National Centre for Atmospheric Science (NCAS). Both images and irradiance measurements were taken at Chilbolton UK Facility for Atmospheric and Radio Research \cite{scienceandtechnologyfacilitiescouncilChilboltonFacilityAtmospheric2003,scienceandtechnologyfacilitiescouncilChilboltonFacilityAtmospheric2016}. The Chilbolton dataset consists of the cloud images and radiometer measurements. A pyranometer collected total global solar irradiance in $W/m^2$ with a temporal resolution of 1 second and about 8000 measurement points per day. Since the images were taken roughly every 5 minutes, the data were pre-processed such that the radiometer data was averaged over a time window of 30 seconds with the resulting value being assigned to one image. The data were aligned based on the timestamps so that the time window for averaging the radiometer data always started at the time stamp of the image.
\cref{fig:raw_RVC} shows the raw measurement data that were available for a single day. As can be seen, the data varies with time of day but shows strong drops in irradiance related to change in cloud conditions. 
To exclude very dark images data points taken between midnight and 3 am data points or with irradiance values below 2 $W/m^2$ were removed from the dataset. Additionally images were removed where objects or animals were blocking the view of the camera and where excessive frost or rain blocked the view. The target data distribution is depicted in \cref{fig:RVC_Hist_Threshold_10_Normal_Images}. 
The data was split into a training and evaluation dataset as well as a separate testing dataset. This split was done by using days 5 to 9 of each month as the fixed testing dataset while using days 15 to 19 for evaluation during training. This left 125000 images from the Chilbolton dataset for training.
To facilitate comparison to other works we also train and test our model on the the NREL-TSI and SIRTA datasets. The NREL-TSI dataset consists of all-sky images taken every 10 minutes \cite{stoffelNRELSolarRadiation1981}. From the NREL-TSI dataset 106000 images taken between 2015 and 2022 were used for training. The SIRTA dataset contains all-sky images captured every 2 minutes at 2 different exposure levels\cite{haeffelinSIRTAGroundbasedAtmospheric2005}. To allow for a direct comparison the data from the SIRTA dataset was filtered and split into train, test and evaluation sets as described in \cite{palettaECLIPSEEnvisioningCLoud2022}. This resulted in 180000 samples being used for model training. The images from all datasets were pre-processed by cropping and resizing them to a size of 224 by 224 pixels. Furthermore, due to the presence of fixed objects in the camera's field of view, a mask of black pixels was applied to the Chilbolton images. All target data were normalised to have a mean of 0 and a standard deviation of 1 using the mean and standard deviation of the training set.

\begin{figure*}[t]
	\centering
	\begin{subfigure}{0.33\linewidth}
	   \centering
 
		\includegraphics[width=0.99\linewidth]{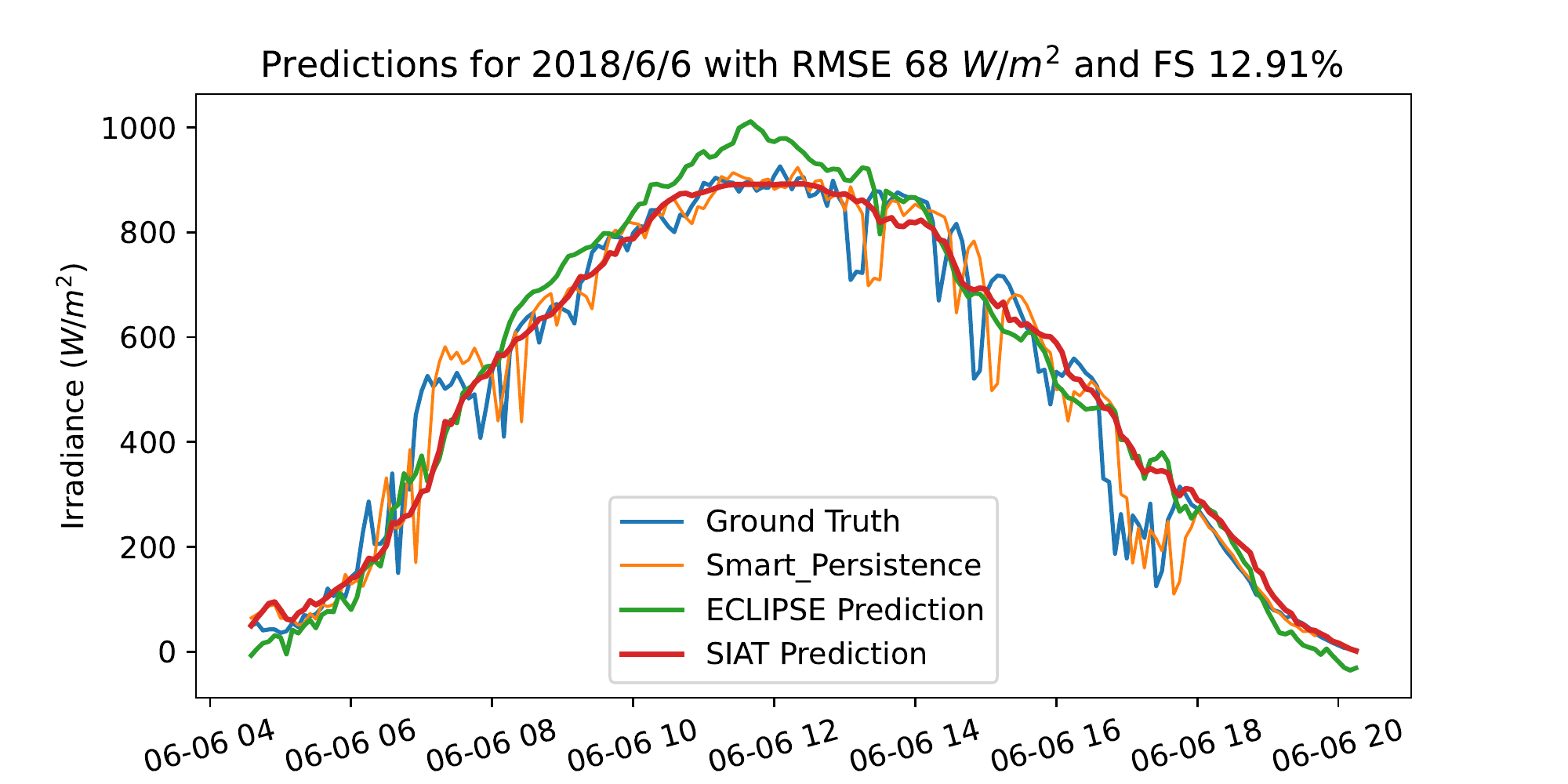}
		\caption{Day with low variability.}
		\label{fig:predictions_PredDay6_Month6_Year2018}
	\end{subfigure}
	\begin{subfigure}{0.33\linewidth}
	   \centering
 
		\includegraphics[width=0.99\linewidth]{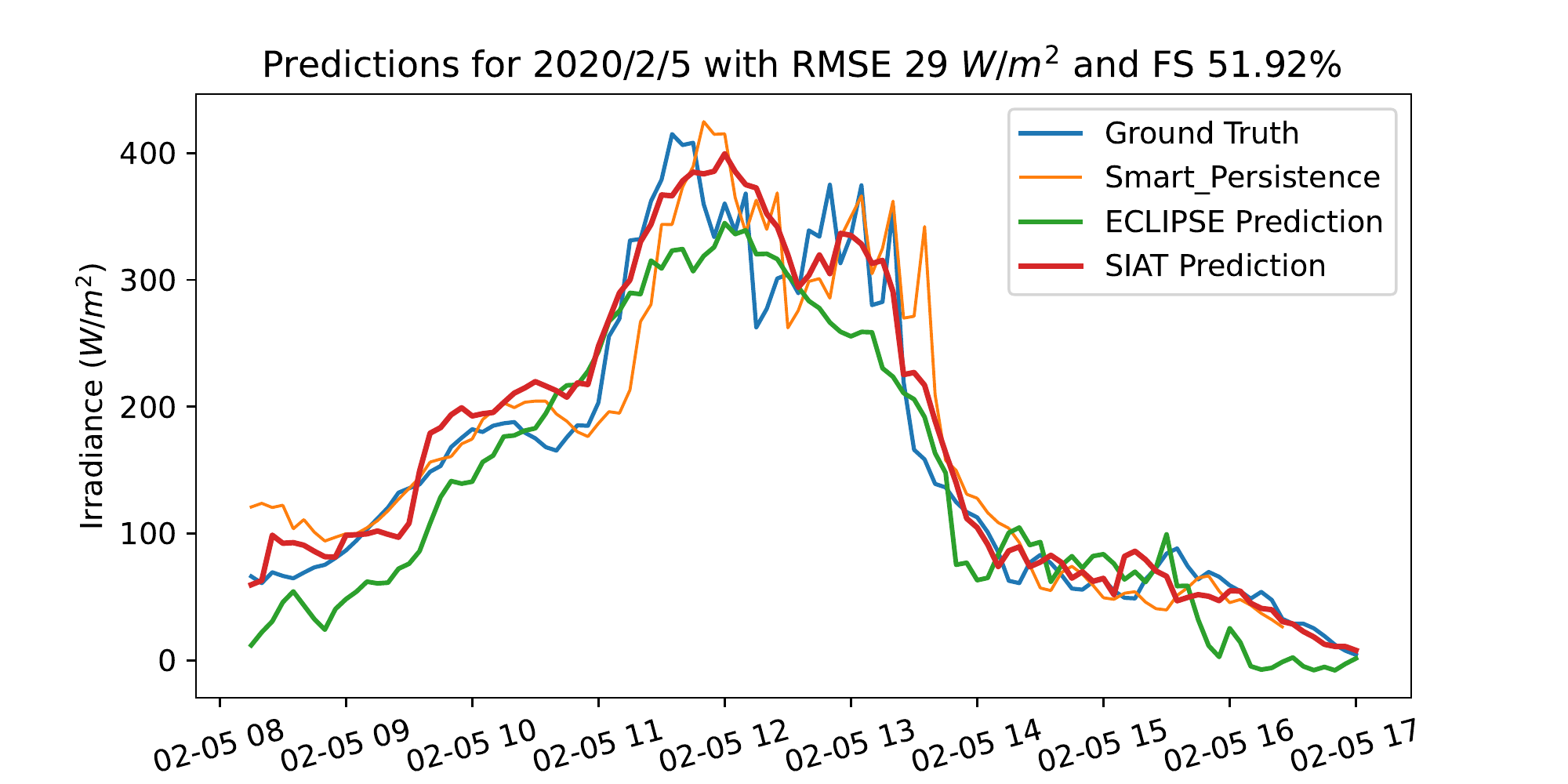}
		\caption{Day with overall low irradiance values.}
		\label{fig:predictions_PredDay5_Month2_Year2020}
	\end{subfigure}
	\begin{subfigure}{0.33\linewidth}
	   \centering
		\includegraphics[width=0.99\linewidth]{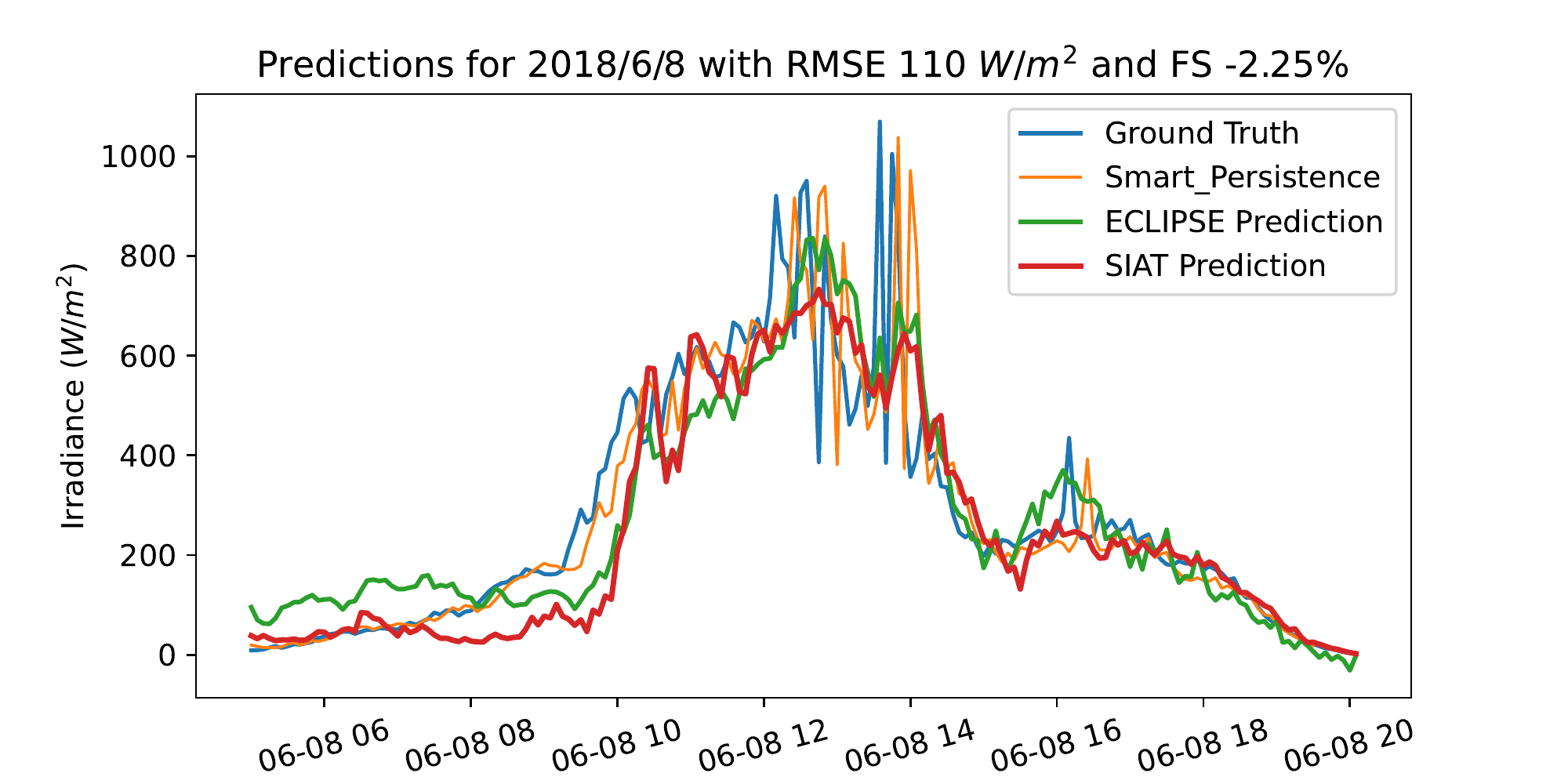}
		\caption{Day with negative FS.}
		\label{fig:predictions_PredDay8_Month6_Year2018}
	\end{subfigure}
	\caption{Comparisons of ground truth irradiance with 15 minute ahead predictions by our SIAT model, the competing ECLIPSE model and the SP approach. The example days are taken from the unseen test set of the Chilbolton dataset.}
	\label{fig:predictions}
\end{figure*}

\section{Implementation}
\label{sec:Implementation}
The backbone of the proposed model consisted of a ViT, that had its weights initialized from a model trained on the ImageNet dataset\cite{rw2019timm,touvronTrainingDataefficientImage2021,dengImageNetLargescaleHierarchical2009}. The backbone was configured to split the 224 by 224 pixel images into 16 by 16 pixel patches which then get flattened and projected to the embedding dimension of 768. The model has a depth of 12 and uses 12 attention heads. To have the model learn to extract task-relevant features, it was trained separately from the full model by having it map single images to the associated irradiance values. To use the backbone in the full model, the final fully connected layer was removed so that the output of the backbone would be the extracted feature vector. 
To use a transfer learning based approach for the backbone on the SIRTA dataset, it was necessary to add an additional 2d-convolutional layer with a kernel size of 3 and a stride of 1 to the model. Since the SIRTA dataset offers two images with different exposures for each irradiance value and the ViT backbone expects images to have only 3 channels, this convolutional layer takes in the channel-concatenated images and projects them to the required channel number.

The head model was configured to use 8 attention heads, to have a depth of 4 and use an internal encoding dimension of 512. Since the time between images varied between 2 and 10 minutes depending on the dataset in question, the prediction horizon varied accordingly. Randomized image augmentation was applied during all training by varying the brightness, contrast, saturation and hue of the images by 1~\% and rotating the images up to 15 degrees. For training the backbone to map images to irradiance values, we use a batch size of 64 and the Adam optimizer with a weight decay of $1\cdot10^{-6}$ and a learning rate of $1\cdot10^{-4}$ and a scheduled cosine anneal being applied to the learning rate every step. The backbone is trained for 11 epochs. For training and testing of the full model, we use a sequence of 5 sequential images to predict 3 timesteps into the future with the reported RMSE being based on the future prediction for the timestep in question.  We use a batch size of 16 and the Adam optimizer with a weight decay of $1\cdot10^{-6}$ and a learning rate of $5\cdot10^{-5}$ with an exponential learning rate warm up and a scheduled cosine anneal being applied to the learning rate every step. The full network is trained for 11 epochs with the backbone staying frozen for 10 epochs. The network is supervised by using intermediate as well as final $L_{irr}$ and $L_{enc}$ loss components. Empirically we found that an equal weighting of all loss components shown in \cref{eq:loss_comp} gives the best performance. All models were implemented using PyTorch\cite{paszkePyTorchImperativeStyle2019} and the code is available at\cite{mercierSIATHttpsGithub2023}. All training was carried out on a machine equipped with a Nvidia RTX 3090 with 24 GB of memory and a i7-7700K with 64 GB of memory.

\section{Results and Discussion}

\begin{table*}[t]
	\centering
	\caption{Comparison of our SIAT model to the competing models for all three datasets. All training scenarios use 5 images as past context. The time between images is 2 minutes for SIRTA, 5 minutes for Chilbolton and 10 minutes for NREL-TSI. The future steps indicates how many timesteps into the future the model predicts. While the results for ECLIPSE model are based on unofficial model implementation, due to code unavailability, with a reported RMSE of 83.8 and 98.5 for 1 and 3 timesteps ahead prediction the results here closely match what the authors report in their publication. In addition to the results we computed ourselves we pull further comparisons for the SIRTA dataset directly from the publication\cite{palettaECLIPSEEnvisioningCLoud2022}.}
            \begin{tabular}{cccccccccc}
            
		\toprule
 \multicolumn{1}{l}{\textbf{}} & \multicolumn{1}{l}{\textbf{}} & \multicolumn{8}{c}{\textbf{Future Steps}}                                                                                                                                                                                                                                                                                                                                                                                                                                                                                   \\ \hline
\multicolumn{1}{l}{}          & \multicolumn{1}{l}{}          & \multicolumn{4}{c}{1}                                                                                                                                                                                                                                        & \multicolumn{4}{c}{3}                                                                                                                                                                                                                                        \\
\textbf{Dataset}              & \textbf{Model}                & \textbf{\begin{tabular}[c]{@{}c@{}}MAE \\ (W/m²)\end{tabular}} & \textbf{\begin{tabular}[c]{@{}c@{}}RMSE\\ (W/m²)\end{tabular}} & \textbf{\begin{tabular}[c]{@{}c@{}}nRMSE\\ (\%)\end{tabular}} & \textbf{\begin{tabular}[c]{@{}c@{}}FS\\ (\%)\end{tabular}} & \textbf{\begin{tabular}[c]{@{}c@{}}MAE \\ (W/m²)\end{tabular}} & \textbf{\begin{tabular}[c]{@{}c@{}}RMSE\\ (W/m²)\end{tabular}} & \textbf{\begin{tabular}[c]{@{}c@{}}nRMSE\\ (\%)\end{tabular}} & \textbf{\begin{tabular}[c]{@{}c@{}}FS\\ (\%)\end{tabular}} \\ \hline
SIRTA                         & Smart Persistence             & \textbf{39.01}                                                 & 93.33                                                          & 25.02                                                         & -                                                          & 62.10                                                          & 129.77                                                         & 34.78                                                         & -                                                          \\
SIRTA                         & SIAT(ours)                    & 42.05                                                          & \textbf{76.94}                                                 & \textbf{20.62}                                                & \textbf{17.57}                                             & \textbf{54.26}                                                 & \textbf{97.60}                                                 & \textbf{26.16}                                                & \textbf{24.79}                                             \\
SIRTA                         & ECLIPSE\cite{palettaECLIPSEEnvisioningCLoud2022}               & 48.94                                                          & 78.90                                                          & 21.15                                                         & 15.46                                                      & 57.61                                                          & 98.64                                                          & 26.44                                                         & 23.99                                                      \\
SIRTA                         & PhyDNet \cite{palettaECLIPSEEnvisioningCLoud2022,guenDeepPhysicalModel2020}                      & -                                                              & 87.70                                                          & 23.51                                                         & 6.00                                                       & -                                                              & 102.00                                                         & 27.34                                                         & 21.10                                                      \\
SIRTA                         & TimeSFormer \cite{palettaECLIPSEEnvisioningCLoud2022,bertasiusSpaceTimeAttentionAll2021}                   & -                                                              & 93.10                                                          & 24.95                                                         & 0.20                                                       & -                                                              & 105.00                                                         & 28.14                                                         & 18.80                                                      \\
SIRTA                         & ConvLSTM\cite{palettaBenchmarkingDeepLearning2021}                      & -                                                              & 95.60                                                          & 25.62                                                         & -2.40                                                      & \textbf{-}                                                     & 107.20                                                         & 28.73                                                         & 17.10                                                      \\ \hline
Chilbolton                    & Smart Persistence             & \textbf{51.96}                                                 & 116.09                                                         & 46.28                                                         & -                                                          & 73.45                                                          & 142.58                                                         & 56.84                                                         & -                                                          \\
Chilbolton                    & SIAT(ours)                    & 57.51                                                          & \textbf{98.12}                                                 & \textbf{39.11}                                                & \textbf{15.48}                                             & \textbf{68.15}                                                 & \textbf{112.00}                                                & \textbf{44.65}                                                & \textbf{21.45}                                             \\
Chilbolton                    & ECLIPSE\cite{palettaECLIPSEEnvisioningCLoud2022}               & 68                                                             & 103.69                                                         & 41.34                                                         & 10.68                                                      & 76.33                                                          & 117.35                                                         & 46.78                                                         & 17.69                                                      \\ \hline
NREL-TSI                      & Smart Persistence             & 88.87                                                          & 169.02                                                         & 43.81                                                         & -                                                          & 151.72                                                         & 241.04                                                         & 62.48                                                         & -                                                          \\
NREL-TSI                      & SIAT(ours)                    & \textbf{66.20}                                                 & 113.71                                                         & 29.47                                                         & 32.73                                                      & \textbf{82.08}                                                 & \textbf{139.71}                                                & \textbf{36.21}                                                & \textbf{42.04}                                             \\
NREL-TSI                      & ECLIPSE\cite{palettaECLIPSEEnvisioningCLoud2022}               & 67.17                                                          & \textbf{112.91}                                                & \textbf{29.27}                                                & \textbf{33.20}                                             & 86.63                                                          & 142.60                                                         & 36.96                                                         & 40.84                                                     
\\

		\bottomrule               
                \end{tabular}
	
	\label{tab:comp}
\end{table*}

We report RMSE, nRMSE, MAE and FS for two different prediction tasks, a single timestep ahead and a three timesteps ahead prediction both using 5 images as the past context. For the newly introduced Chilbolton dataset our SIAT model achieves an RMSE of 112 $W/m^2$ for the 15 minute or three timesteps ahead prediction. For the same data the SP reference model achieves an RMSE of 142.58 $W/m^2$, this corresponds to an FS of 21.45~\%. \cref{tab:comp} gives on overview on how our model performs on different datasets and compared with competing models. As can be seen our model outperforms the competing model on all datasets for the three timestep ahead prediction setting while for the single timestep ahead prediction, we outperform the competing model only on the SIRTA and Chilbolton datasets. While ECLIPSE shows slightly higher FS for this dataset and scenario, the MAE of our model is still better. 
It is notable that for the very short term prediction of a single timestep for the Chilbolton and SIRTA datasets, the MAE for the SP approach is lower than that of both the ECLIPSE and our SIAT model. The much higher MAE for the SP model on the single timestep ahead prediction for the NREL-TSI can be expected since the time between images is double that of the Chilbolton dataset. Due to the code not being officially available for the competing ECLIPSE model, the presented results are based on an unofficial implementation. For the SIRTA dataset the authors of the ECLIPSE model report an RMSE of 83.8 and 98.5 $W/m^2$ for the one and three timestep predictions, respectively. Using the unofficial implementation we find the RMSE to be 78.9 and 98.64 $W/m^2$ for the same forecasting scenarios. Since these results either beat or match the ones the authors report themselves the implementation can be considered faithful and the computed results for the other two datasets valid. Notably, we also outperform the models for which the results in \cref{tab:comp} were pulled directly from literature. The authors did not report MAE. We attribute the high performance of our model to the transformers ability to use the temporal and spatial information contained in the series of images used for the prediction.

Comparisons of predictions for the 15 minute ahead case to ground truth for three randomly selected days from the unseen test set of the Chilbolton dataset are shown in \cref{fig:predictions}. On the example data shown in \cref{fig:predictions_PredDay6_Month6_Year2018}, the model outperforms the SP reference model with a FS of 12.91~\% for a day with relatively low variability. Here it can also be seen that unlike the ECLIPSE model, our SIAT model avoids the overestimation of the peak irradiance. However, both models fail to predict the sharp rise between 6 and 8am and the dip between 4 and 6 pm.

\cref{fig:predictions_PredDay5_Month2_Year2020} shows an example day where the model does very well, reaching a FS of 51.92~\%. It successfully predicts the rise in irradiance around 11 am as well as the sharp drop between 1 and 2 pm. While the ECLIPSE model also anticipates the rise, it overall suffers from underestimating the irradiance values as well as anticipating changes that do not occur, as seen in the anticipated dip in irradiance between 3 and 4 pm.

\cref{fig:predictions_PredDay8_Month6_Year2018} shows an example day with low irradiance values for most of the day with sharp peaks between noon and 2 pm. While the SIAT model follows the overall shape of the curve it is still results in a negative FS. In this particular case the SIAT model underestimates the irradiance up to 10 am while the ECLIPSE model initially overestimates them. Both models are unable to predict the rise in irradiance between 9 and 10am. As a general observation both the ECLIPSE model and our SIAT model give predictions that result in an overall smoother curve than the ground truth irradiance, however the SIAT model does a better job of predicting changes and of avoiding large over- and underpredictions.

\begin{figure}[h]
	\centering
	\begin{subfigure}{0.32\linewidth}
	   \centering
		\includegraphics[width=0.99\linewidth]{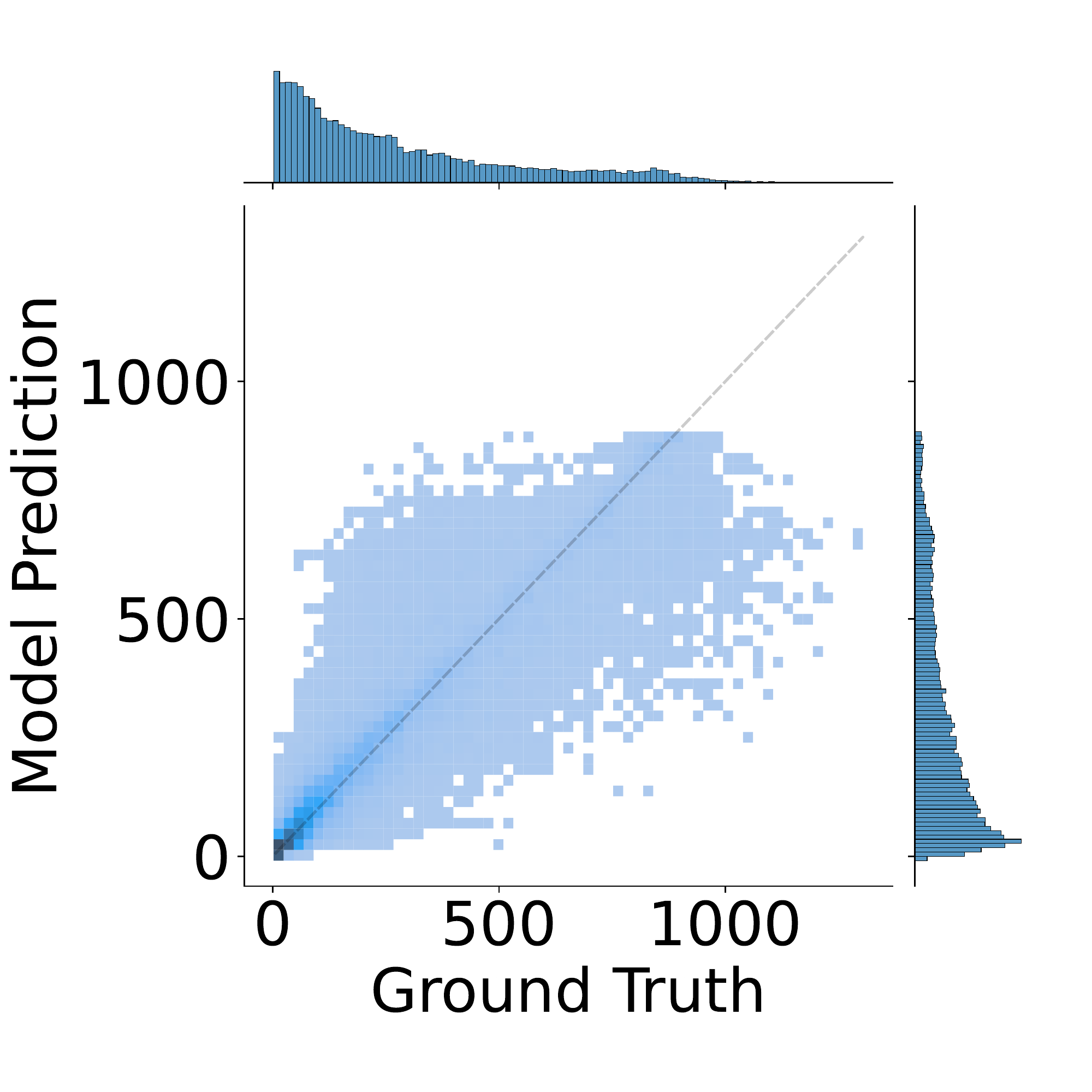}
	\caption{Model predictions.}
	\label{fig:predictions_scatter}
	\end{subfigure}
	\begin{subfigure}{0.32\linewidth}
	   \centering
		\includegraphics[width=0.99\linewidth]{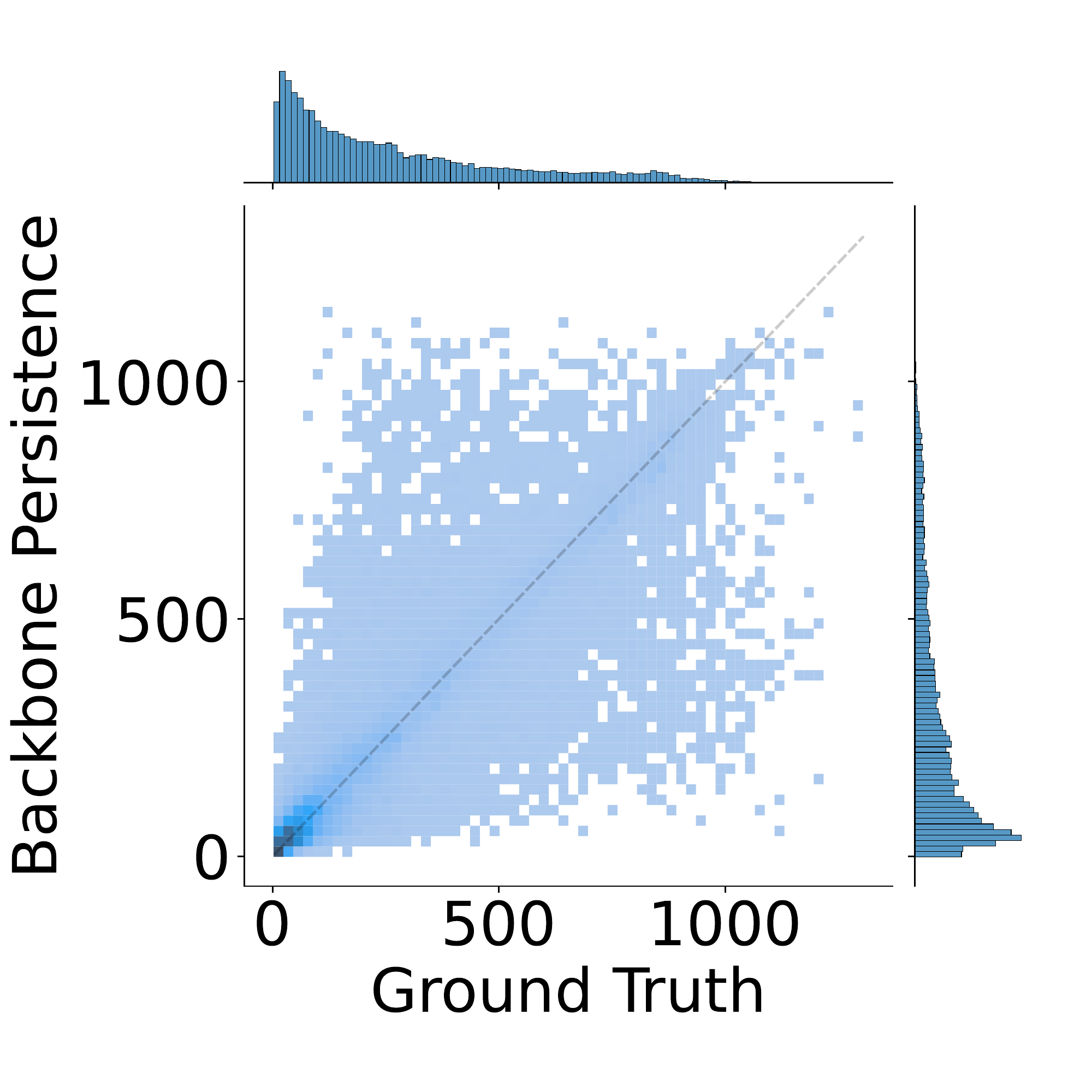}
		\caption{BP predictions.}
		\label{fig:backbone_persistence}
	\end{subfigure}
	\begin{subfigure}{0.32\linewidth}
	   \centering
		\includegraphics[width=0.99\linewidth]{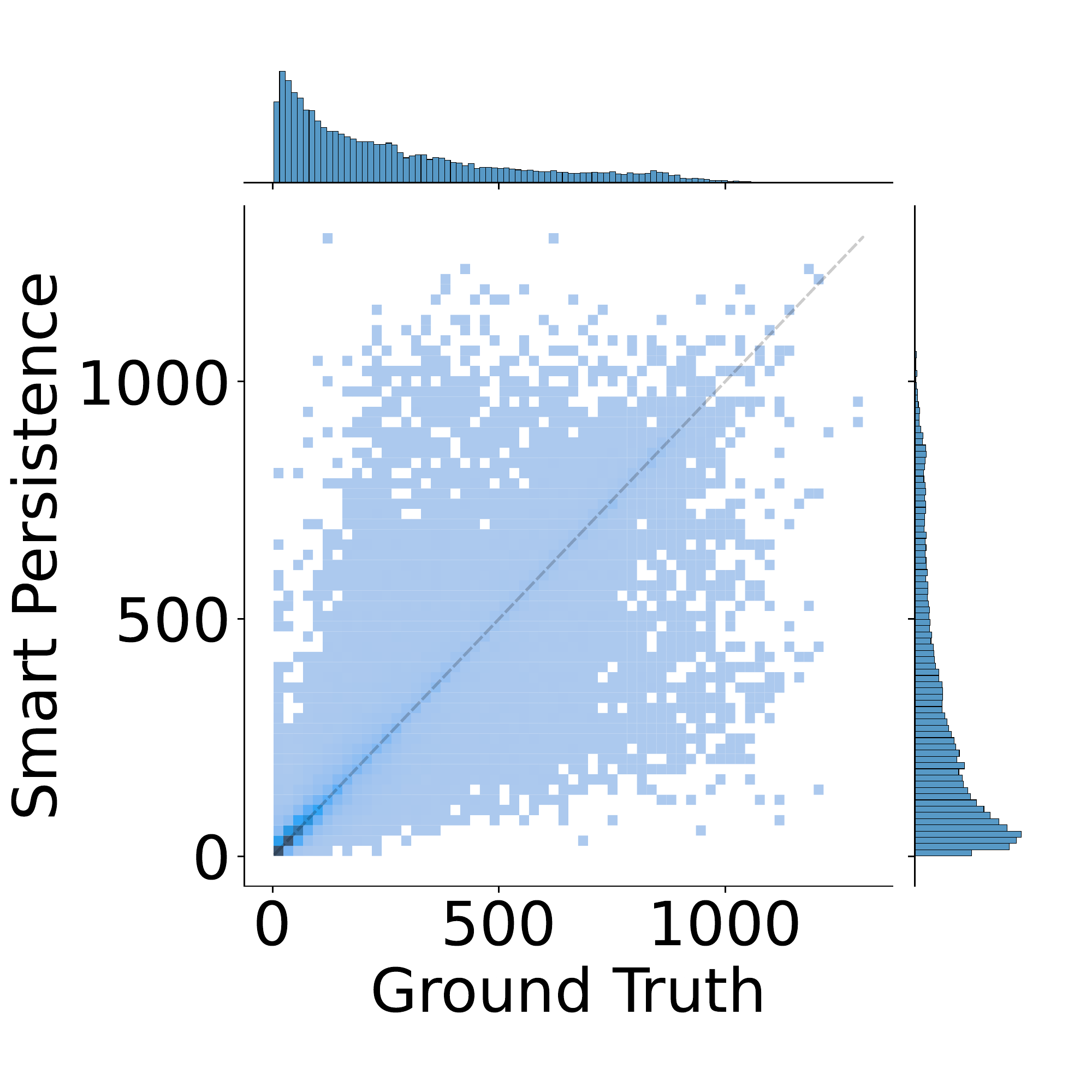}
		\caption{SP predictions.}
		\label{fig:Hist_joint_Chilbolton_Smart_Persistence}
	\end{subfigure}
	\caption{Density plots comparing predicted irradiance values to ground truth with the grey dashed line representing the ideal case. We show predictions for the SIAT model as well as for both persistence approaches based predictions for 15 minute ahead forecasting task on the Chilbolton dataset. Darker blue indicates higher density. Both persistence approaches yield a much broader distribution compared to our SIAT model.}
	\label{fig:smart_persistence}
\end{figure}

A comparison of predicted irradiance values and ground truth for all samples in the Chilbolton test set is shown in \cref{fig:predictions_scatter}. The two-dimensional histogram shows that there is no significant bias in the model's predictions. However, the model avoids predictions of very high irradiance values above 900 $W/m^2$, since these are rare in the data and as can be seen in the examples days in \cref{fig:predictions}, our SIAT model does sometimes underestimate the irradiance values, especially for very brief peaks in the irradiance curves.

\begin{figure}[t]
	\centering
	\begin{subfigure}{0.24\linewidth}
		\centering
		\includegraphics[width=\linewidth]{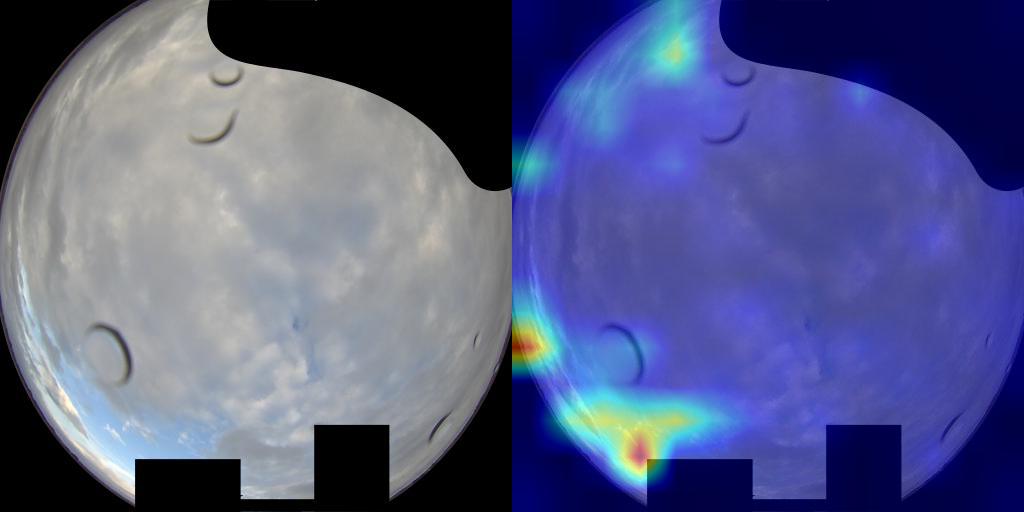}
	\end{subfigure}
	\begin{subfigure}{0.24\linewidth}
		\centering
		\includegraphics[width=\linewidth]{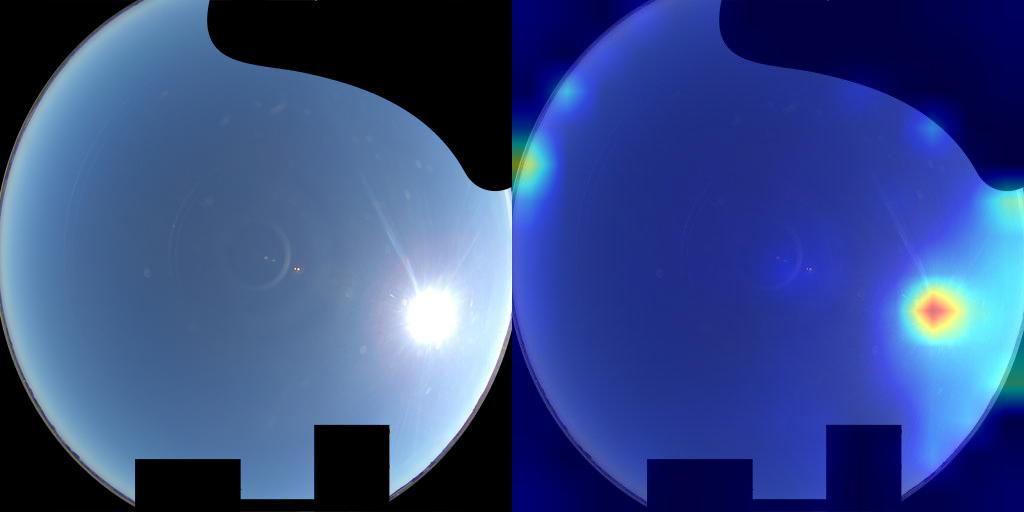}
	\end{subfigure}	
	\begin{subfigure}{0.24\linewidth}
		\centering
		\includegraphics[width=\linewidth]{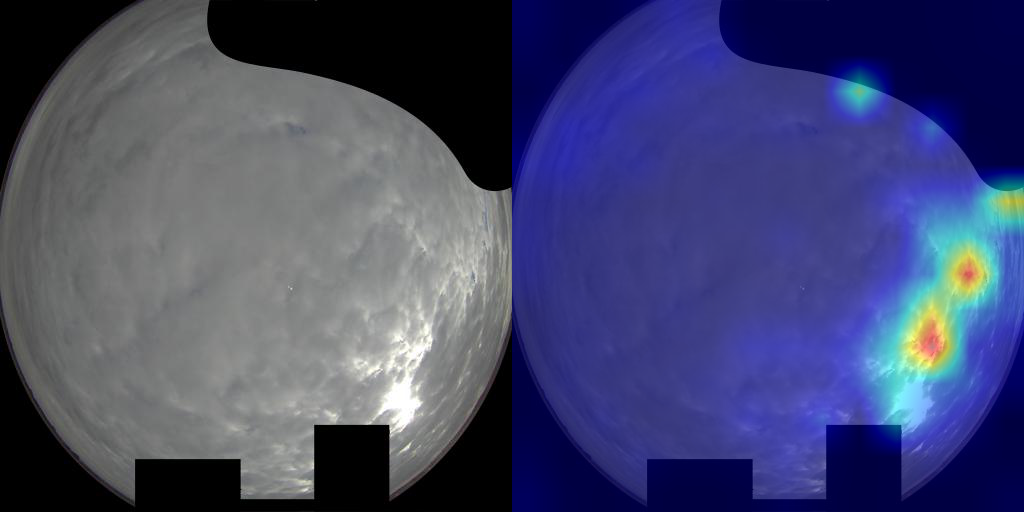}
	\end{subfigure}
	\begin{subfigure}{0.24\linewidth}
		\centering
		\includegraphics[width=\linewidth]{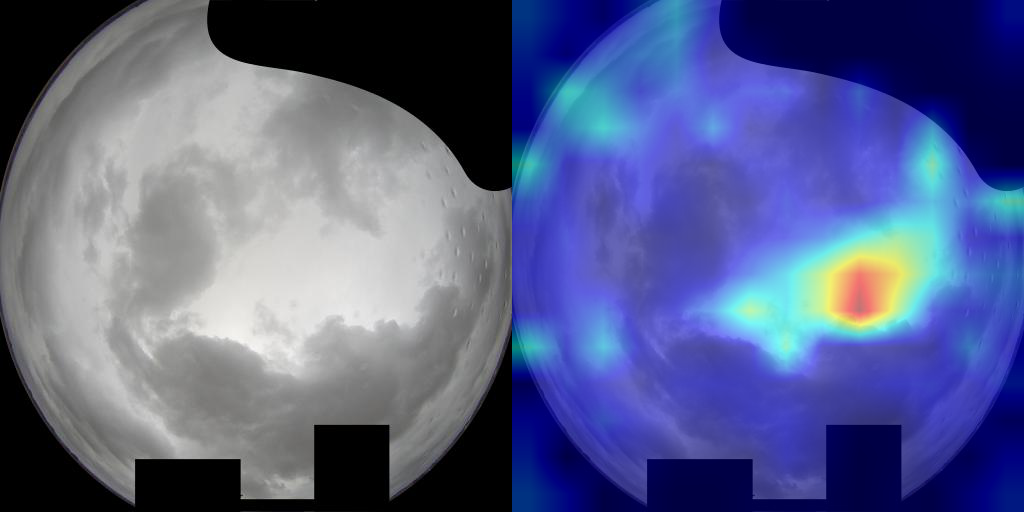}
	\end{subfigure}
	\begin{subfigure}{0.24\linewidth}
		\centering
		\includegraphics[width=\linewidth]{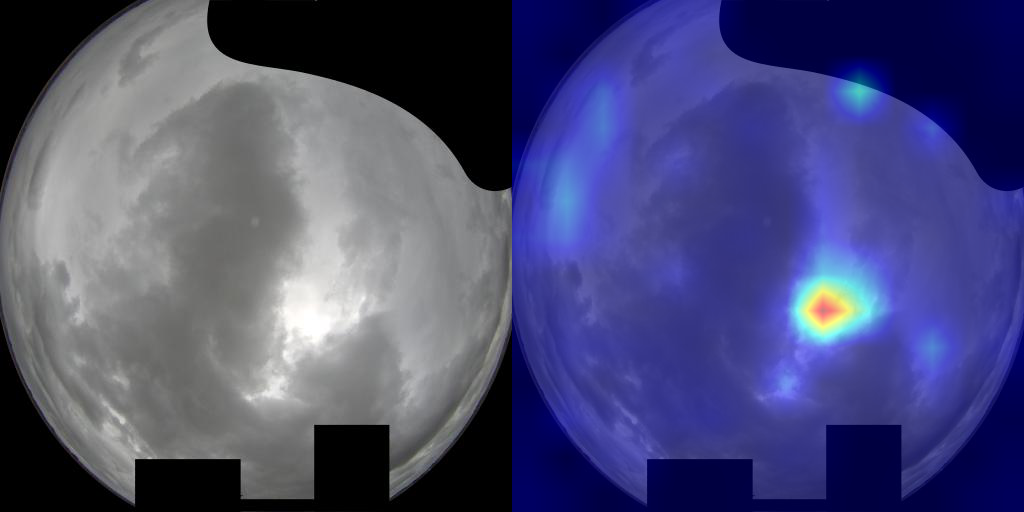}
	\end{subfigure}
	\begin{subfigure}{0.24\linewidth}
		\centering
		\includegraphics[width=\linewidth]{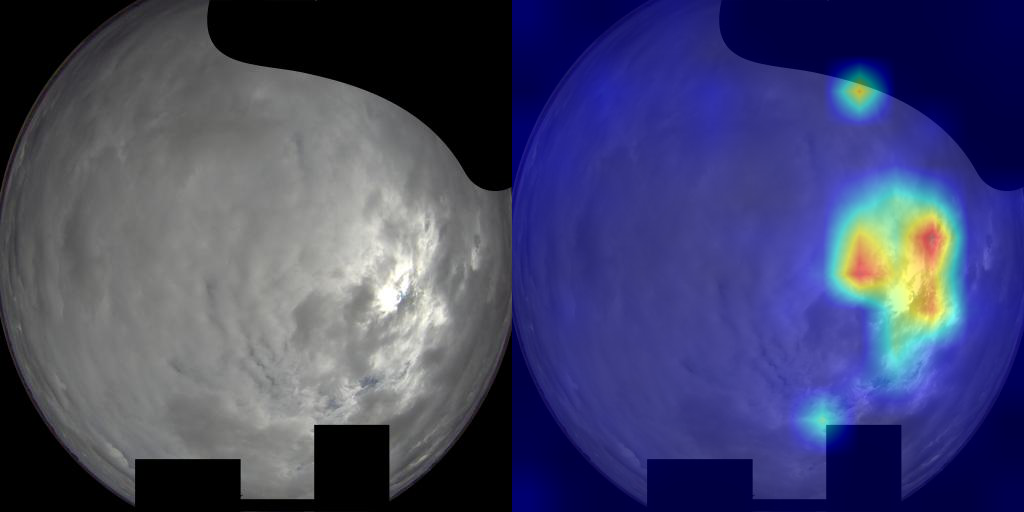}
	\end{subfigure}
	\begin{subfigure}{0.24\linewidth}
		\centering
		\includegraphics[width=\linewidth]{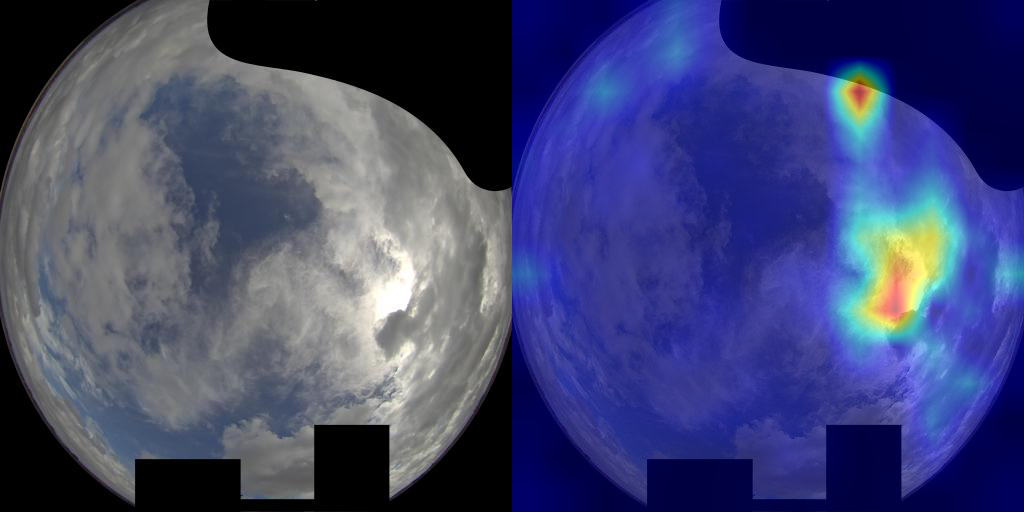}
	\end{subfigure} 
	\begin{subfigure}{0.24\linewidth}
		\centering
		\includegraphics[width=\linewidth]{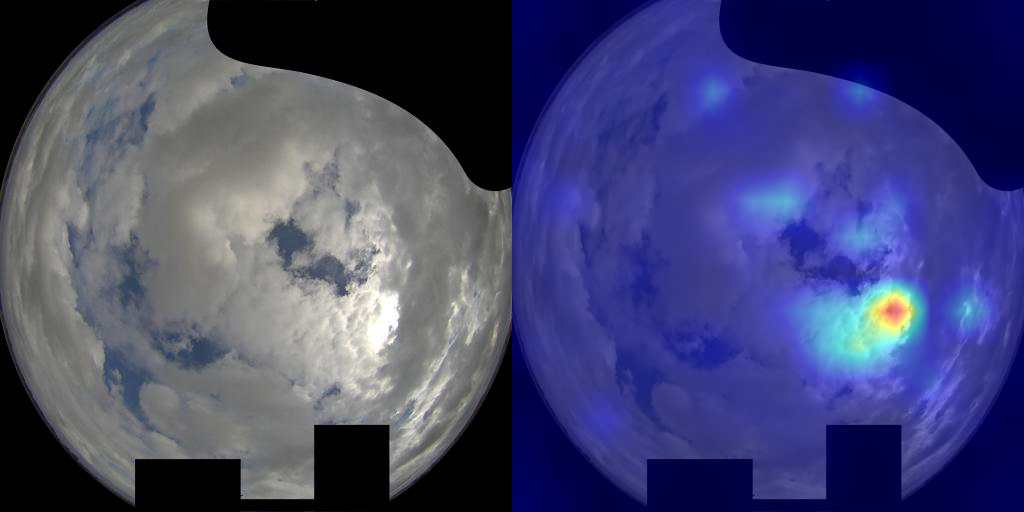}
	\end{subfigure}
	\begin{subfigure}{0.24\linewidth}
		\centering
		\includegraphics[width=\linewidth]{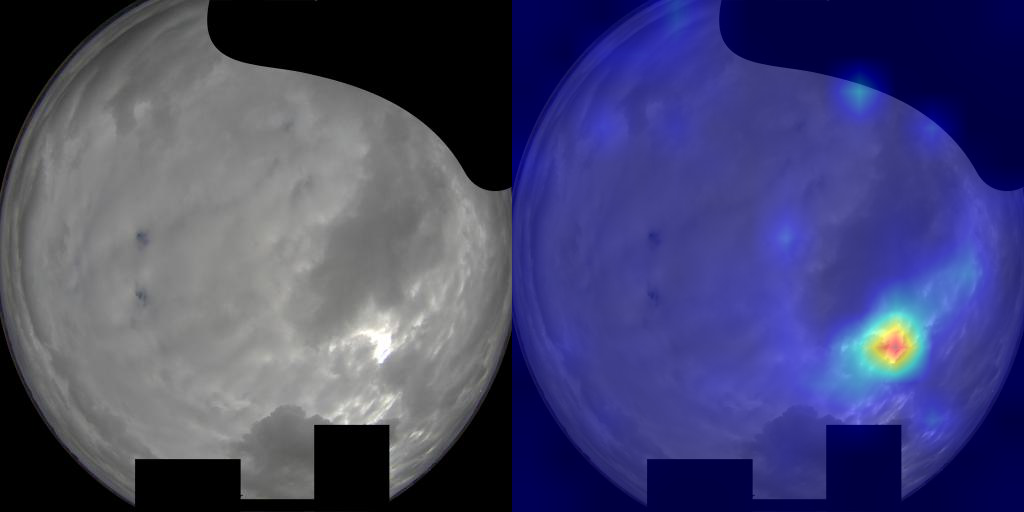}
	\end{subfigure}
	\begin{subfigure}{0.24\linewidth}
		\centering
		\includegraphics[width=\linewidth]{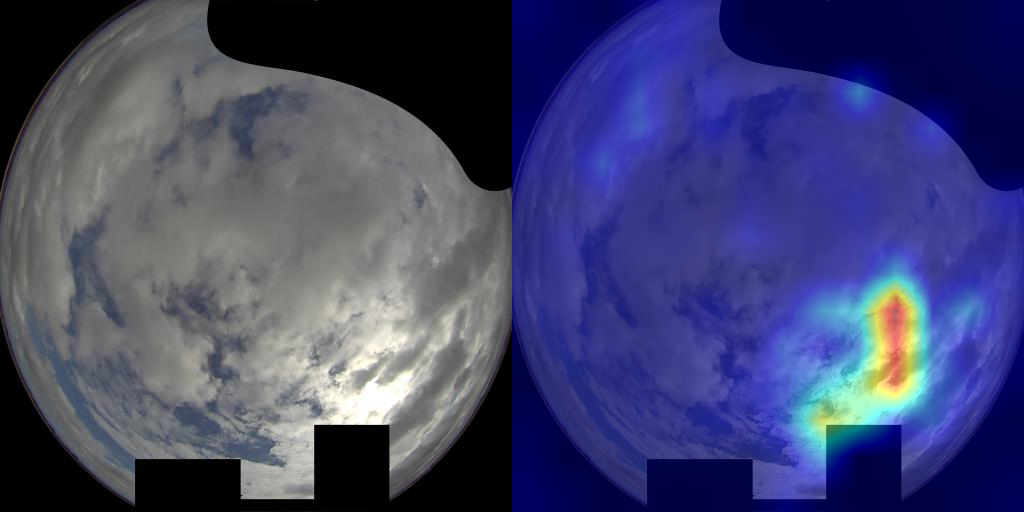}
	\end{subfigure}
	\begin{subfigure}{0.24\linewidth}
		\centering
		\includegraphics[width=\linewidth]{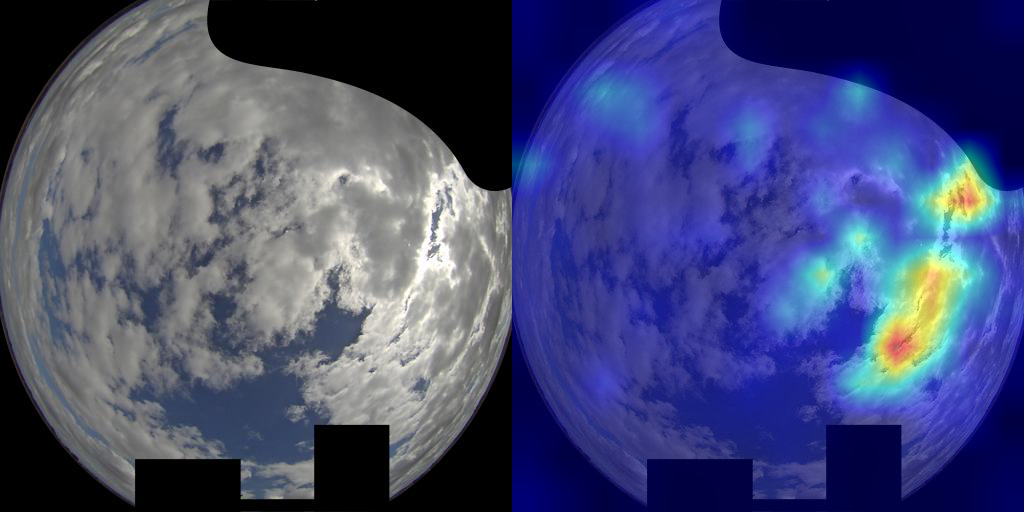}
	\end{subfigure}
	\begin{subfigure}{0.24\linewidth}
		\centering
		\includegraphics[width=\linewidth]{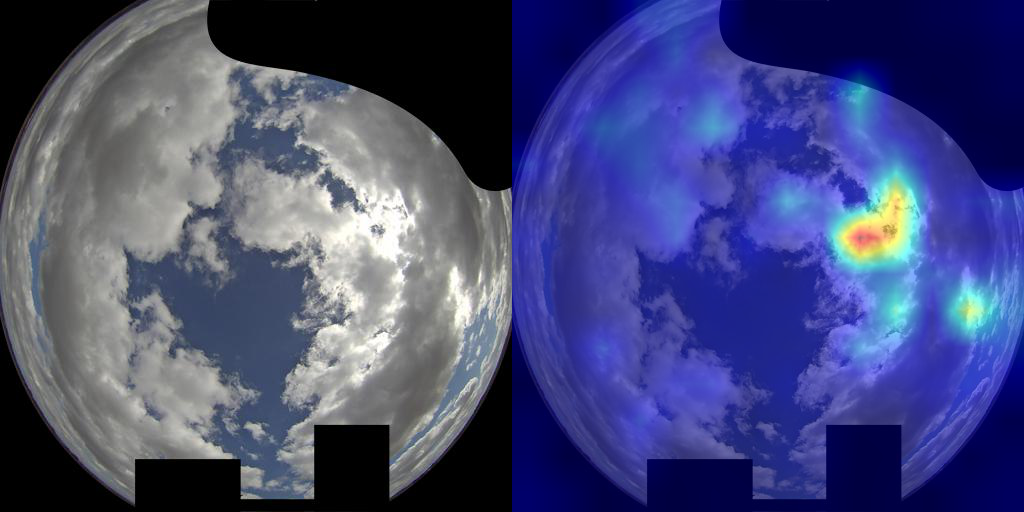}
	\end{subfigure}
	\begin{subfigure}{0.24\linewidth}
		\centering
		\includegraphics[width=\linewidth]{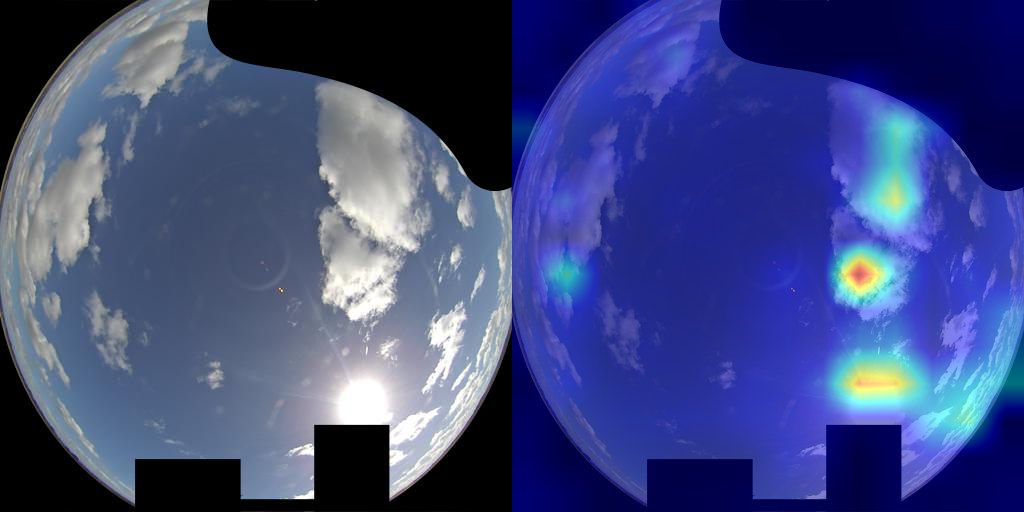}
	\end{subfigure}
	\begin{subfigure}{0.24\linewidth}
		\centering
		\includegraphics[width=\linewidth]{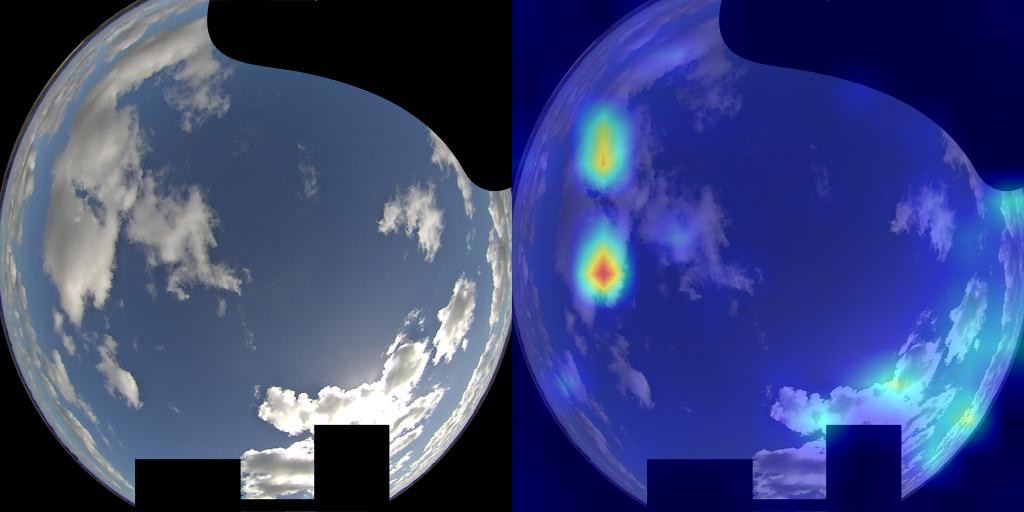}
	\end{subfigure} 
	\begin{subfigure}{0.24\linewidth}
		\centering
		\includegraphics[width=\linewidth]{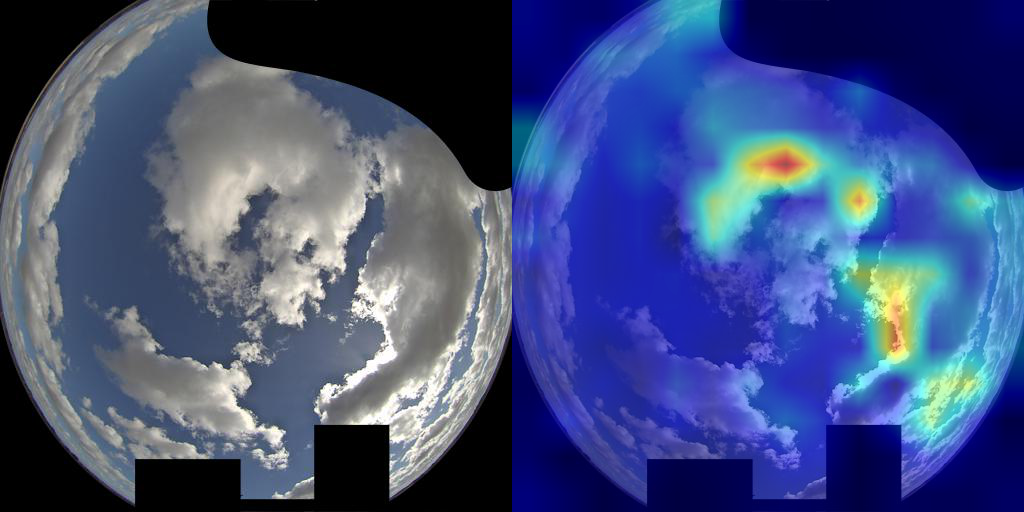}
	\end{subfigure}
	\begin{subfigure}{0.24\linewidth}
		\centering
		\includegraphics[width=\linewidth]{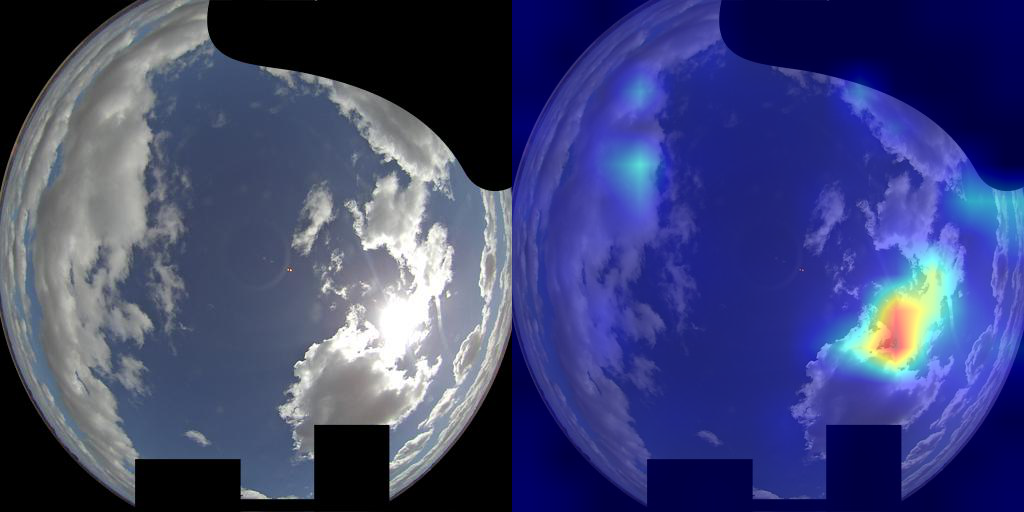}
	\end{subfigure}
	\caption{Attention maps for the fine-tuned backbone model for a variety of sky conditions. The overlaid heatmap visualises the areas of the image that the model learns to pay attention to with red signifying higher attention activation and blue signifying low activation. It can be seen that the fine-tuned backbone learns to attend to the areas surrounding the sun and nearby cloud formations. All attention maps were produced using attention rollout\cite{abnarQuantifyingAttentionFlow2020}.}
	\label{fig:attention_maps}
\end{figure}

We show that the feature encoder model successfully learns to attend to irradiance relevant features of the sky images as can be seen in the example attention maps shown in \cref{fig:attention_maps}. The backbone's attention predominantly falls on the sun as well as cloud formations near it as these are the most important features for making irradiance predictions. Furthermore, the attention maps show that the model is able to extract relevant features under a variety of sky conditions from clear skies with the sun clearly visible to strongly overcast days with the sun barely shining through. The attention maps were produced using attention rollout\cite{abnarQuantifyingAttentionFlow2020}.

In the following paragraphs we show the results of a range of ablation studies analyzing the SIAT architecture using the 15 minute ahead forecasting task on the Chilbolton dataset.
To evaluate by how much the GPT-2 based decoder outperforms a simple persistence prediction, we use the ViT backbone with a densely connected layer on top to map each image in the Chilbolton testset to an irradiance value (as is done in the backbone training stage) and shift this value by three timesteps. We refer to this as the backbone persistence (BP) approach. This results in an overall RMSE of 139.58 $W/m^2$ and an MAE of 79.98 $W/m^2$. Since our full model achieves an RMSE of 112 $W/m^2$ and an MAE of 72.35 $W/m^2$, this clearly demonstrates that the GPT-2 based decoder performs much better than a simple persistence model. This is also borne out in \cref{fig:backbone_persistence} as the BP approach results in a large spread around the ideal.

\begin{table}[h]
	\centering
	\caption{Comparison of training the model with a convolution based backbone, a ResNet152, and a transformer based backbone\cite{heDeepResidualLearning2015}. Evaluation metrics are reported for 15 minute ahead irradiance prediction using the Chilbolton dataset.}
	\small
	\begin{tabular}{@{}llll@{}}
		\toprule
		\textbf{Backbone} & \textbf{RMSE ($W/m^2$)} & \textbf{MAE ($W/m^2$)}     & \textbf{FS (\%)}     \\
		\midrule
		ResNet152 & 114.28 & 73.03 & 19.85\\ 
		ViT & \textbf{112} & \textbf{68.15} & \textbf{21.45}\\
		\bottomrule
	\end{tabular}
	\label{tab:backbones}
\end{table}

To gauge the effect that the attention based backbone has on the overall performance of the model, we replaced the ViT backbone with a ResNet152\cite{heDeepResidualLearning2015,rw2019timm}, with the comparison being shown in \cref{tab:backbones}. The three stage training procedure was kept the same with the backbone being trained separately. For the 15 minute ahead forecasting task the model using the ViT based backbone performed better than the ResNet152 on all evaluation metrics, with the FS dropping from 21.45 to 19.85~\%.

\begin{table}[h]
	\centering
	\caption{Comparison of model performance with and without the first stage of training where the backbone gets trained to map an irradiance value to a single image. Evaluation metrics are reported for 15 minute ahead irradiance prediction using the Chilbolton dataset.}
 
	\resizebox{0.99\linewidth}{!}{
	\small
	\begin{tabular}{@{}llll@{}}
		\toprule
		\textbf{Training stages} & \textbf{RMSE ($W/m^2$)} & \textbf{MAE ($W/m^2$)}     & \textbf{FS (\%)}     \\
		\midrule
		2 & 113 & 71.75 & 20.75\\ 
		3 & \textbf{112} & \textbf{68.15} & \textbf{21.45}\\
		\bottomrule
	\end{tabular}
    }
	\label{tab:backbones_sep_training}
\end{table}

Since we utilize a three stage training process where the backbone is first trained to map an all-sky image to an irradiance value with the linear head network then being removed to allow the backbone to act as a feature extractor for the all-sky images, we also ran the training of the full model without this first stage of training with the results shown in \cref{tab:backbones_sep_training}. As can be seen utilizing a three stage training procedure rather than a two stage procedure boosts performance on all evaluation metrics with the model's FS increasing from 20.75 to 21.45~\%.

\begin{table}[h]
	\centering
	\caption{Comparison of results of training the model using either MAE or MSE as the supervision loss function. All results shown are for the 15 minute ahead forecasting task for the Chilbolton dataset.}
	\small
	\begin{tabular}{@{}llll@{}}
		\toprule
		\textbf{Training Loss}  &  \textbf{RMSE ($W/m^2$)} & \textbf{MAE ($W/m^2$)}     & \textbf{FS (\%)}\\
		\midrule
		MAE & 116.02 & \textbf{66.41} & 18.63 \\
		MSE & \textbf{112} & 68.15 & \textbf{21.45} \\ 
		\bottomrule
	\end{tabular}
	\label{tab:loss_methods}
\end{table}
\cref{tab:loss_methods} shows how the model performs when the MAE loss function is used to supervise the model during training. As can be seen the performance as measured by the overall RMSE and FS suffers with the FS falling from 21.45~\% to 18.63~\%, however the overall MAE sees some improvement.

\begin{table}[h]
\centering
\caption{Evaluation results of supervising the SIAT model using different loss components. All results shown are for the 15 minute ahead forecasting task for the Chilbolton dataset.}
	\small
	\resizebox{0.99\linewidth}{!}{
	\begin{tabular}{@{}llll@{}}
            \toprule
            \textbf{Loss components}   &  \textbf{RMSE ($W/m^2$)} & \textbf{MAE ($W/m^2$)}     & \textbf{FS (\%)}  \\
            		\midrule
            $L_{irr,f}$                                 & 113.8 & 70.1 & 20.19 \\
            $L_{irr,f}$ + $L_{enc}$                       & 113.16 & 72.35 & 20.64 \\
            $L_{irr,f}$ + $L_{irr,i}$          & 113.65 & \textbf{69.68} & 20.29 \\
            $L_{irr,f}$ + $L_{irr,i}$ + $L_{enc}$ & \textbf{112} & 72.35 & \textbf{21.45} \\
            \bottomrule
        \end{tabular}
        }
\label{tab:lossMech}
\end{table}
As \cref{tab:lossMech} shows, including the encoding loss component in the supervision of the model brings the largest improvement in FS. Notably, including the encoding loss component results in worsening of the MAE metric. While only using the intermediate and final irradiance loss components results in the lowest MAE, the FS is significantly worse.

\begin{figure}[h]
    \centering
    \includegraphics[width=0.6\linewidth]{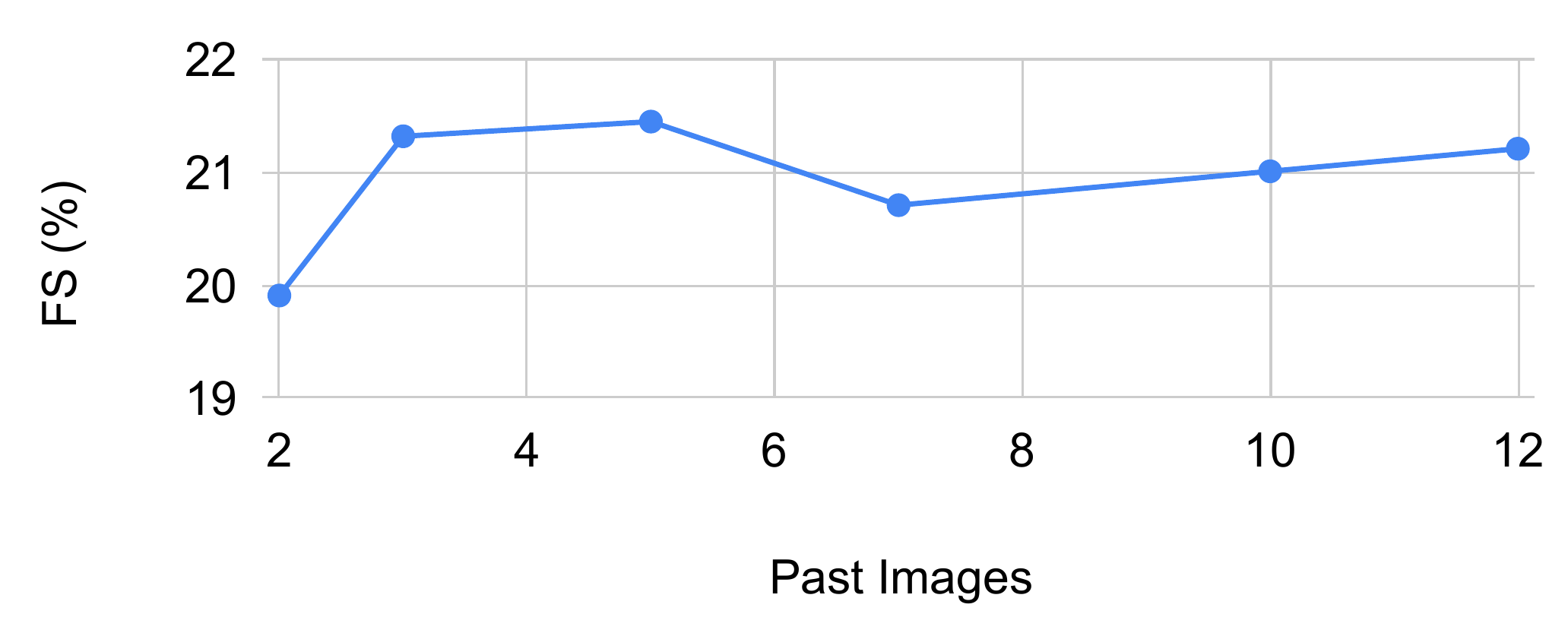}
    \caption{Evaluation results for varying past context lengths while keeping the number of predicted timesteps fixed at 3. Results shown for the Chilbolton dataset.}
    \label{fig:seq_len_fs}
\end{figure}
\cref{fig:seq_len_fs} shows how the FS changes when different number of images are used as past context. For the 15 minute ahead prediction task on the Chilbolton dataset a past context of 5 images is found to be ideal.

\section{Conclusion}

We present SIAT, a transformer based framework for the task of forecasting solar irradiance using a sequence of all-sky images without the use of auxiliary data. A ViT backbone serves as a feature extractor to create a feature vector for each frame in the sequence. Our approach then utilizes the temporal relationship contained in the extracted features via a GPT-2 based decoder network. Our training scheme first has the backbone learn to map images to irradiance values to ensure the backbone learns to extract task relevant features. This backbone remains frozen for the first part of the training of the full model. We supervise the model by both its ability to predict future features as well as irradiance values. In the last stage of training the backbone is unfrozen to allow for further fine-tuning of the full architecture. We show that the model successfully learns to attend to important features in the sky images. For the 15 minute ahead forecasting task achieve an RMSE of 112 $W/m^2$ on the Chilbolton dataset, which corresponded to an FS of 21.45~\%. For the three timestep prediction we demonstrate that SIAT outperforms competing models on all datasets. 

{\small
\bibliographystyle{ieee_fullname}
\bibliography{Solar_Predict}
}

\end{document}